\documentclass{article}

\PassOptionsToPackage{numbers, compress}{natbib}
\usepackage[preprint]{neurips_2025}

\usepackage[utf8]{inputenc} 
\usepackage[T1]{fontenc}    
\usepackage{hyperref}       
\usepackage{url}            
\usepackage{booktabs}       
\usepackage{amsfonts}       
\usepackage{nicefrac}       
\usepackage{microtype}      
\usepackage{xcolor}         

\usepackage{amsmath}
\usepackage{graphicx}
\usepackage{wrapfig}

\title{T2SMark: Balancing Robustness and Diversity in Noise-as-Watermark for Diffusion Models}

\author{%
  Jindong Yang$^{1,2}$,  Han Fang$^{3*}$, Weiming Zhang$^{1,2}$, Nenghai Yu$^{1,2}$, Kejiang Chen$^{1,2}$\thanks{Corresponding authors.}\\
  $^1$University of Science and Technology of China \\
  $^2$Anhui Province Key Laboratory of Digital Security\\
  $^3$National University of Singapore \\
  \texttt{dx929@mail.ustc.edu.cn, fanghan@nus.edu.sg} \\
  \texttt{\{ynh, zhangwm, chenkj\}@ustc.edu.cn}
}

\begin{document}

\maketitle

\begin{abstract}
Diffusion models have advanced rapidly in recent years, producing high-fidelity images while raising concerns about intellectual property protection and the misuse of generative AI. Image watermarking for diffusion models, particularly Noise-as-Watermark (NaW) methods, encode watermark as specific standard Gaussian noise vector for image generation, embedding the infomation seamlessly while maintaining image quality. For detection, the generation process is inverted to recover the initial noise vector containing the watermark before extraction. However, existing NaW methods struggle to balance watermark robustness with generation diversity. Some methods achieve strong robustness by heavily constraining initial noise sampling, which degrades user experience, while others preserve diversity but prove too fragile for real-world deployment. 
To address this issue, we propose T2SMark, a two-stage watermarking scheme based on Tail-Truncated Sampling (TTS). 
Unlike prior methods that simply map bits to positive or negative values, TTS enhances robustness by embedding bits exclusively in the reliable tail regions while randomly sampling the central zone to preserve the latent distribution. Our two-stage framework then ensures sampling diversity by integrating a randomly generated session key into both encryption pipelines. 
We evaluate T2SMark on diffusion models with both U-Net and DiT backbones. Extensive experiments show that it achieves an optimal balance between robustness and diversity. Our code is available at \href{https://github.com/0xD009/T2SMark}{https://github.com/0xD009/T2SMark}.
\end{abstract}

\section{Introduction}
\label{sec:intro}

\begin{wrapfigure}[12]{r}{0.3\textwidth}
    \centering
    \includegraphics[width=1\linewidth]{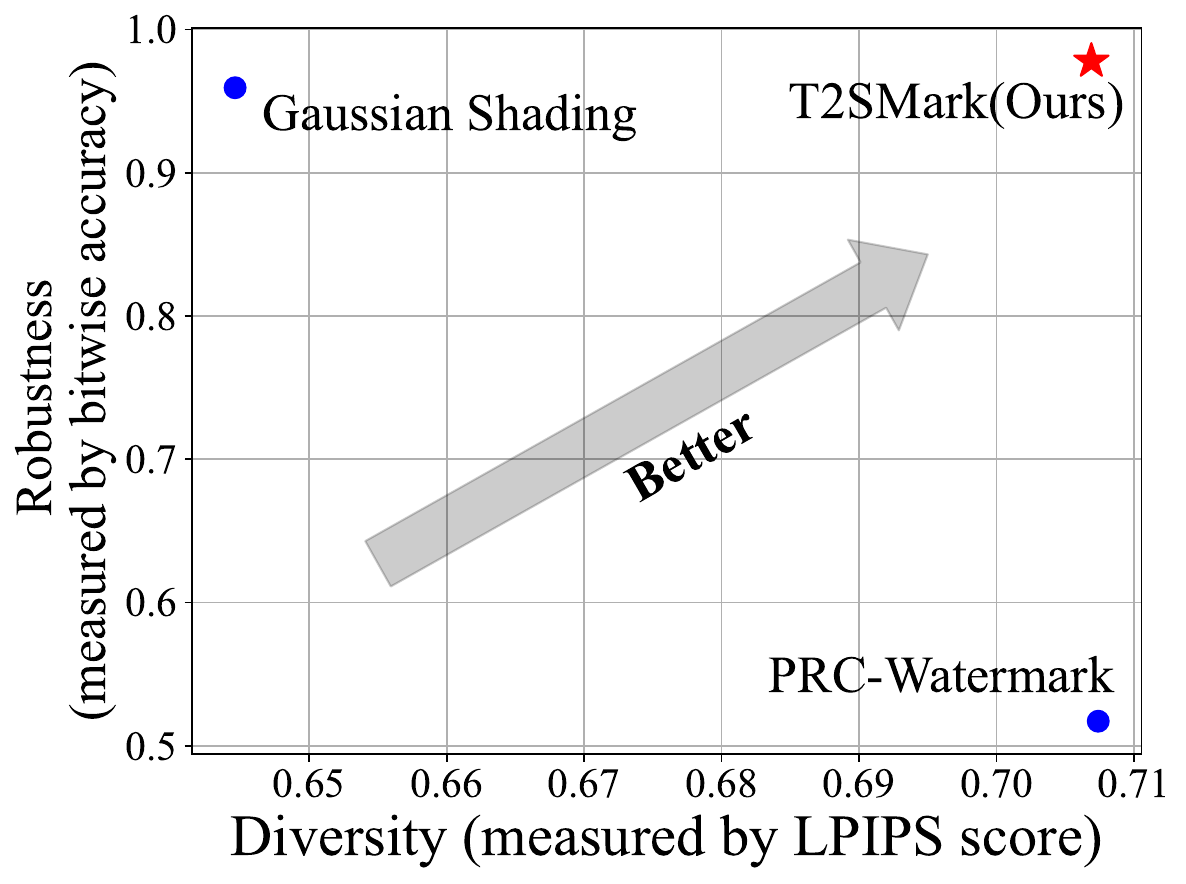}
    \caption{T2SMark strikes an optimal balance between robustness and diversity.}
\end{wrapfigure}

In recent years, diffusion models \cite{ddpm_2020,ddim_2021,ldm_2022,dit_2023} have achieved groundbreaking advances in generative AI, especially in image synthesis. Owing to their iterative denoising process, these models can produce highly realistic images. As their architecture scales and training datasets expand, diffusion models continually improve in both generation quality and controllability, driving rapid progress in AIGC technologies.

However, this swift evolution also brings new challenges: the authenticity of generated content is increasingly difficult to verify, and diffusion models have become potent instruments for spreading disinformation and manipulating public opinion. For example, during the 2024 U.S. presidential election, U.S. officials accused Russia and other countries of using generative AI to fabricate fake news, images, and videos to influence voter sentiment and disrupt the electoral process\footnote{\href{https://www.npr.org/2024/09/23/nx-s1-5123927/russia-artificial-intelligence-election}{U.S. officials say Russia is embracing AI for its election influence efforts}. NPR, September 23, 2024.}. Meanwhile, service providers invest vast amounts of data and computing power to train these models and urgently need effective mechanisms to safeguard their intellectual property against unauthorized use. Together, these concerns have increased the demand for robust solutions to protect the copyright and trace the provenance of AI-generated images, making this a focal point of current academic research.

To address these challenges, watermarking presents a fundamental solution and has shown considerable promise. Among various techniques, Noise-as-Watermark (NaW) methods, including Gaussian Shading (GS) \cite{gs_2024} and PRC-Watermark (PRCW) \cite{prcw_2025}, stand out. At its core, NaW maps watermark codewords onto a standard Gaussian noise vector, uses it as the initial noise for image generation, and thus embeds watermark information while preserving the noise distribution. For extraction, one simply inverts the diffusion process to recover the embedded noise and subsequently decodes it to retrieve the watermark. The reversible bijection offered by the forward and backward diffusion processes between high-entropy Gaussian vectors and the low-entropy image manifold allows NaW methods to achieve high-capacity, lossless embedding for individual images without the need for supplementary training.

The embedding process in most Noise-as-Watermark (NaW) methods comprises two core steps: discrete encoding and continuous sampling. 
The former step converts the watermark message into codewords, while the latter transforms those discrete binary codes into continuous noise.
The challenge in discrete encoding lies in balancing the need for randomness, which makes the watermark difficult to predict, with the need for robustness against various distortions –- goals that are often contradictory. Continuous sampling, on the other hand, is critical for maintaining imperceptibility by precisely preserving the initial Gaussian noise distribution. Current methods make different trade-offs: GS \cite{gs_2024} employs a simple repetition code for high robustness but lacks randomness, depending entirely on the sampling process for variation. PRCW \cite{prcw_2025}, while using a pseudorandom error-correcting code \cite{prc_2024} for better undetectability, exhibits weak robustness and can fail when subjected to common inversion errors.

\begin{wrapfigure}[10]{r}{0.4\textwidth}
    \centering
    \includegraphics[width=1\linewidth, trim=0cm 0.25cm 0cm 0cm, clip]{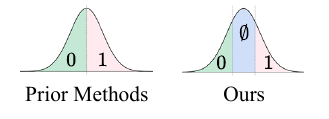}
    \caption{Tail-Truncated Sampling divides the distribution into three regions: bit-0, bit-1 and an undecided zone (\(\emptyset\)).}
    \label{fig:tts}
\end{wrapfigure}

To overcome these limitations and achieve a balance of both robustness and diversity, we propose \textbf{T2SMark}, \textit{a novel two-stage watermarking scheme built upon Tail-Truncated Sampling (TTS)}. While previous efforts have concentrated primarily on discrete encoding strategies, we observed that Gaussian samples closer to the origin are significantly more vulnerable to sign errors when noise is present. This finding inspired our TTS approach: instead of a simple positive/negative mapping for bits, TTS divides the Gaussian distribution into three distinct areas: one for bit-0, one for bit-1, and an undecided region (see Figure \ref{fig:tts}). Information is embedded only within the more reliable tail regions, with the central region being sampled randomly to compensate for diversity and maintain the overall distribution. 
Additionally, our two-stage framework introduces controlled randomness via hierarchical key encryption. A static master key encrypts a randomly generated session key in the first stage, and this session key is then used to encrypt the actual watermark bits in the second stage. This layered encryption randomizes both parts of the watermark codewords, thereby further ensuring the generation diversity. Finally, we improve detection and decoding by employing multidimensional projections of the reconstructed Gaussian noise, allowing us to fully leverage its rich continuous information.

Evaluated on diffusion models \cite{ldm_2022, sd3_2024} with both U-Net \cite{unet_2015} and DiT \cite{dit_2023} backbones, T2SMark demonstrates an optimal trade-off between watermark robustness and generative diversity.

\section{Related Work}

\subsection{Diffusion Models}

Diffusion models \cite{ddpm_2020,ddim_2021,ldm_2022,dit_2023} iteratively corrupt data by injecting Gaussian noise and train a denoising network to reverse this process. The canonical DDPM framework \cite{ddpm_2020} employs a stochastic Markov chain for multi-step sampling, delivering outstanding sample quality at the cost of slow inference. Building on DDPM, DDIM \cite{ddim_2021} reformulates the random process as a deterministic ODE, transforming sampling into an invertible mapping that both accelerates generation and enables inversion—making it the backbone of many inversion-based watermarking techniques \cite{gs_2024, prcw_2025, treering_2023}. Moreover, latent diffusion models (LDMs) \cite{ldm_2022} first compress high-dimensional data into a lower-dimensional latent space and then perform noise addition and denoising in that compact space. This strategy dramatically reduces computation while preserving generation quality and significantly boosting efficiency.

Inspired by the Transformer’s success in NLP \cite{transformer2017, bert_2019, gpt_2020}, researchers have introduced self-attention into diffusion models. Diffusion transformers (DiTs) \cite{dit_2023} leverage the scalability of transformer architectures for image generation. Today, many leading diffusion pipelines \cite{sd3_2024, sora_2024, pixart_2023} adopt DiT backbones, driving advances in high-fidelity and efficient synthesis.

Generation diversity is an essential metric for assessing a model’s creative capacity. ODE-based samplers \cite{ddim_2021, dpm_2022,  dpmpp2022} achieve faster inference by discarding stochastic noise during denoising, which means that diversity depends entirely on the initial noise vector. As a result, any watermarking scheme that alters or constrains the initial noise sampling can have a direct effect on generation diversity.

\subsection{Image Watermarking for Diffusion Models}

Image watermarking methods for diffusion models can be divided into three main categories. 1)\textbf{ Post‐processing schemes} \cite{dwt2007, rivagan2019, mbrs_2021, cin_2022} embed a watermark directly into the generated image—using either traditional transform‐domain techniques \cite{dwt2007} or deep learning–based methods \cite{rivagan2019, mbrs_2021, cin_2022}—without modifying the sampling process. While this preserves model diversity, it inevitably degrades visual quality and provides only limited robustness. 2) \textbf{Fine‐tuning schemes} \cite{stablesig_2023, wadiff_2024} inject watermarks by adapting either the diffusion model’s denoiser or, in a latent diffusion setup, the VAE \cite{vae_2023} backbone; these methods maintain sample fidelity but require costly weight updates as model architectures grow. 3) \textbf{Inversion‐based schemes} \cite{gs_2024, prcw_2025, treering_2023} embed information into the initial Gaussian noise sample and recover it by inverting the diffusion process. This category includes perturbation‐based approaches such as Tree-Ring \cite{treering_2023} and RingID \cite{ringid_2024}, which adds robust Fourier-domain patterns at the expense of distributional bias, and Noise-as-Watermark (NaW) techniques such as Gaussian Shading (GS) \cite{gs_2024} and PRC-Watermark (PRCW) \cite{prcw_2025}, which use distribution-preserving sampling to achieve high embedding capacity without shifting the noise distribution. Our T2SMark method also falls into this NaW paradigm. Building upon the principles of Tree-Ring, other methods such as ROBIN \cite{robin_2024} and ZoDiac \cite{zodiac_2024} focus on ownership detection rather than message embedding. These are 0-bit watermarking schemes, designed solely to verify if an image was generated by a specific model. They leverage gradient-based optimization to embed an imperceptible signature, achieving impressive fidelity and competitive robustness. However, their design philosophy differs significantly from our approach. By not needing to be strictly identical to an unwatermarked original, training-free NaW methods can fully leverage the initial noise space to achieve a high capacity (e.g., 256-bit) for traceability. Given these fundamental differences in goals (detection vs. traceability) and methodology (optimization-based vs. training-free), we do not conduct a direct comparison with them in this paper.

\section{Method}

\subsection{Threat Model}

\textbf{Scenarios.} This study considers four key parties: the provider, who operates a closed API image generation service maintaining proprietary model weights and training data; the compliant user, who adheres to platform policies and utilizes the outputs legitimately; the unauthorized user, who illicitly obtains images to claim them as their own; and the malicious user, who exploits the API to generate and disseminate harmful or deceptive content.

\textbf{Detection.} The provider embeds a single-bit watermark into every image generated and serves through the API. Both compliant and malicious users receive only watermarked outputs, whereas unauthorized users have no means to obtain unmarked images. Even after typical manipulations, such as compression or cropping, our extraction algorithm robustly recovers the embedded watermark bit, thereby furnishing incontrovertible proof of the provider’s legitimate copyright while simultaneously signaling the image’s synthetic origin.

\textbf{Traceability.} Each API account is uniquely assigned an identity watermark. In instances of misuse, extracting the embedded identity watermark from a suspect image (potentially after data augmentation designed to evade tracing) enables the provider to match it against the account database, thus pinpointing the source, deterring unauthorized distribution, and holding malicious actors accountable.

\subsection{Overview}

\begin{figure}[h]
    \centering
    \includegraphics[width=1\linewidth]{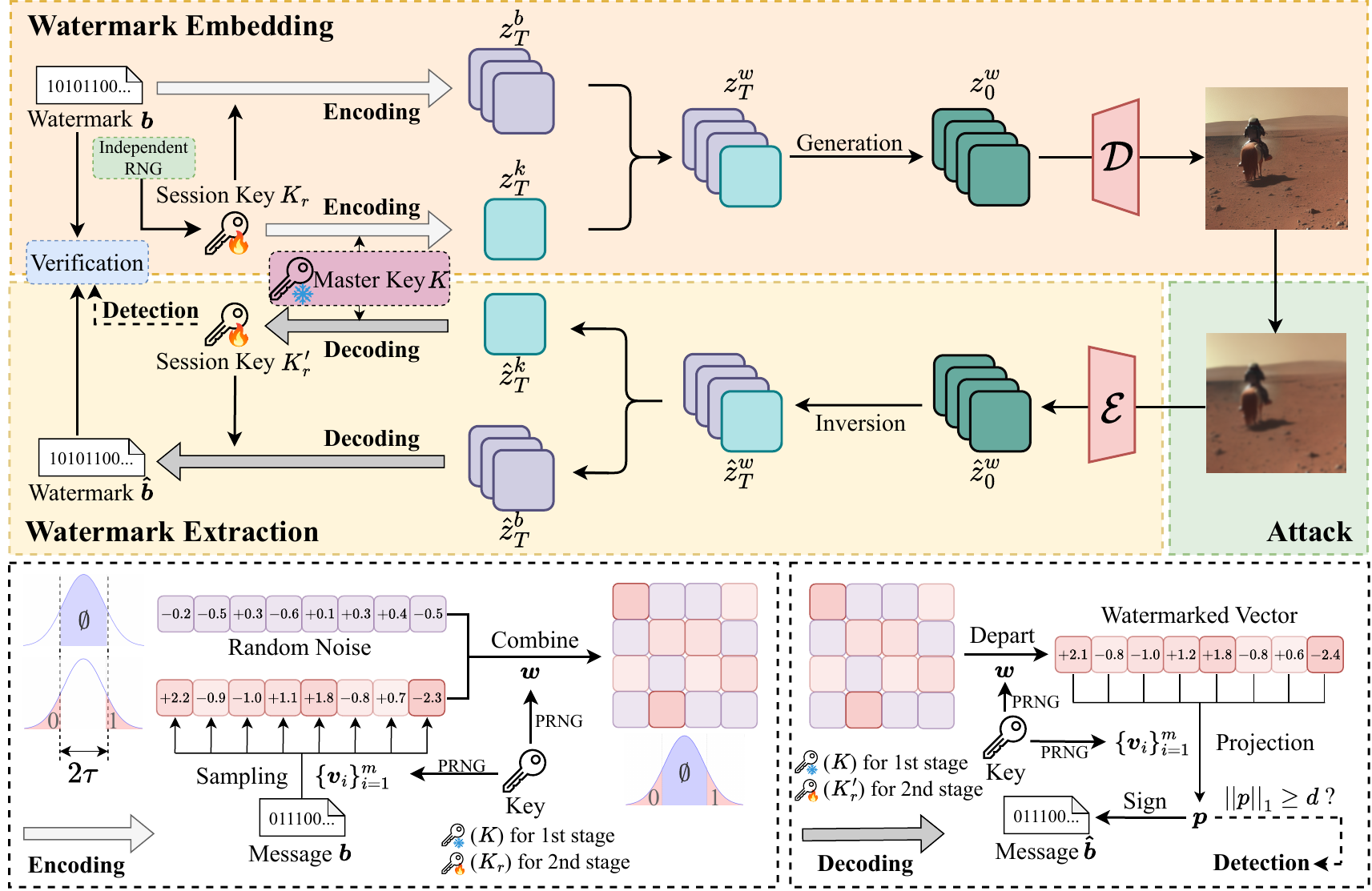}
    \caption{The framework of T2SMark.
    Building on the NaW paradigm, Tail-Truncated Sampling enhances robustness while the two-stage structure introduces controlled randomness.}
    \label{fig:framework}
\end{figure}

T2SMark can be clearly illustrated by Fig.~\ref{fig:framework}.
We first split the noise vector into two segments: the first encodes a random session key under a fixed master key, and the second uses that session key to encode the actual watermark bits. Because both segments depend on the random key, the full vector remains randomized.

Formally, in the single-stage setting, we treat the \(n\)-dimensional noise space as a direct sum of orthogonal subspaces. The key pseudorandomly selects a hyperplane through the origin that splits each subspace into two half spaces for binary encoding.
During sampling, we employ \emph{Tail-Truncated Sampling}: points are drawn along each hyperplane’s normal and kept at least a distance \(\tau\) from the boundary. This thresholding carves out a robust watermark-encoding subspace of larger-magnitude vectors that resist cross-boundary perturbations, while the remaining dimensions carrying no bits retain randomness.

At extraction, we reconstruct the noise via an inversion algorithm \cite{ddim_2021} and project it onto the same set of hyperplane normals. Each bit is recovered from the sign of its projection, which identifies the half-space containing the sample, while detection uses the projection’s $L_1$ norm as a confidence measure.
This approach fully exploits the continuous structure of the noise vector. The detailed scheme follows in subsequent sections.

\subsection{Watermark Encoding}

Here, \textit{encoding} means mapping the watermark into a continuous noise vector, not just discrete coding.
Let \(n\in\mathbb{N}\) be the initial noise dimension, \(\tau\ge0\) the truncation threshold for Tail-Truncated Sampling, \(m\in\mathbb{N}\) the watermark length, and \(\boldsymbol{b}\in\{\pm1\}^m\) the bit-vector.

Based on the threshold $\tau$, we determine the expected number of tail-sampled dimensions $k$ and the per-bit subspace dimension $r$:
\begin{align}
k = 2\,\Phi(-\tau)\,n,
\qquad
r = \bigl\lfloor k / m \bigr\rfloor,
\end{align}
Using a secret key $K$ as the seed for a PRNG (Pseudorandom Number Generator), we generate a set of $m$ pseudorandom vectors $\{\boldsymbol{v}_j\}_{j=1}^m$, where each $\boldsymbol{v}_j \in \{-1,0,+1\}^n$. These vectors are designed with orthogonal, non-overlapping supports and serve as the normal vectors for the bit-encoding hyperplanes within the watermark embedding subspace:
\begin{align}
\|\boldsymbol{v}_j\|_0 = r,\quad
\mathrm{supp}(\boldsymbol{v}_j)\cap\mathrm{supp}(\boldsymbol{v}_{j'}) = \emptyset\quad(j\neq j').
\end{align}
Next, we form the binary mask $\boldsymbol{w} = \sum_{j=1}^m \lvert\boldsymbol{v}_j\rvert \in\{0,1\}^n$. This mask partitions the $n$ dimensions into the watermark-encoding subspace ($w_i=1$) and the random-noise subspace ($w_i=0$). Tail-Truncated Sampling (TTS) is applied to sample components for the noise vector $\boldsymbol{z}$:
\begin{align}
z_i &\sim
\begin{cases}
\mathcal{TN}(0,1;[-\tau,\tau]), & w_i=0,\\
\mathcal{TN}(0,1;(-\infty,-\tau]\cup[\tau,\infty)), & w_i=1.
\end{cases}
\end{align}
where \(\mathcal{TN}(\mu,\sigma^2;I)\) denotes the normalized truncation of \(\mathcal{N}(\mu,\sigma^2)\) over interval \(I\). 
The final watermarked noise vector $\boldsymbol{z}^w$ is then constructed by combining the signs corresponding to the bits in the tail-sampled dimensions with their magnitudes from $\boldsymbol{z}$, while incorporating the random samples from the central dimensions. This is achieved by:
\begin{align}
\boldsymbol{z}^w = \boldsymbol{w}\odot\lvert\boldsymbol{z}\rvert\odot \sum_{j=1}^m b_j\,\boldsymbol{v}_j + (\mathbf{1}^n-\boldsymbol{w})\odot\boldsymbol{z},
\end{align}
where $\odot$ is the Hadamard product. The term $\sum_{j=1}^m b_j\,\boldsymbol{v}_j$ represents the directional encoding of the bits $\boldsymbol{b}$ within the subspace defined by $\{\boldsymbol{v}_j\}_{j=1}^m$.

\subsection{Watermark Decoding}
Upon reconstructing the watermarked noise $\widehat{\boldsymbol{z}}^w$ via diffusion inversion \cite{ddim_2021}, we regenerate the set of vectors $\{\boldsymbol{v}_j\}_{j=1}^m$ using the same secret key $K$. The projection vector $\boldsymbol{p} \in \mathbb{R}^m$ is computed by taking the dot product of $\widehat{\boldsymbol{z}}^w$ with each vector $\boldsymbol{v}_j$:
\begin{align}
p_j = \langle\widehat{\boldsymbol{z}}^w,\boldsymbol{v}_j\rangle, \quad \text{for } j=1,\dots,m.
\end{align}
The watermark bit-vector $\widehat{\boldsymbol{b}}$ is recovered directly from the signs of the projection vector components:
\begin{align}
\widehat{\boldsymbol{b}} = \mathrm{sign}(\boldsymbol{p}),
\end{align}
where $\mathrm{sign}(\cdot)$ is applied elementwise to the $m$-dimensional vector $\boldsymbol{p}$. With a per-bit encoding dimension \(r\) less than \(\lfloor n / m \rfloor\) (indicating reduced repetition), a higher signal-to-noise ratio (SNR) is achieved by TTS, theoretically leading to an even lower probability of bit error under the AWGN assumption (see Appendix \ref{sec:ber}).

\subsection{Two-Stage Watermark}

We split the \(n\)-dimensional noise into two segments of dimensions \(n_k\) and \(n_b\) (with \(n_k + n_b = n\)), often corresponding to channelwise partitioning. First, using the master key \(K\), we sample an \(n_k\)-dimensional vector \(\boldsymbol{z}^k\) that encodes a random session key \(K_r\). Next, with \(K_r\), we sample an \(n_b\)-dimensional vector \(\boldsymbol{z}^b\) containing the watermark bits \(\boldsymbol{b}\). The combined watermarked noise is given by \(\boldsymbol{z}^w=\boldsymbol{z}^k \| \boldsymbol{z}^b\), where \((\cdot\|\cdot)\) denotes vector concatenation.

During extraction, we partition the reconstructed noise \(\widehat{\boldsymbol{z}}^w\) into \(\widehat{\boldsymbol{z}}^k\) and \(\widehat{\boldsymbol{z}}^b\). We first recover \(K_r'\) from \(\widehat{\boldsymbol{z}}^k\) via the master key \(K\), and then retrieve the watermark \(\widehat{\boldsymbol{b}}\) from \(\widehat{\boldsymbol{z}}^b\) using \(K_r'\). 
The session key \(K_r\) thus serves a dual role: embedded payload in the first segment and key for the second, injecting randomness across \(\boldsymbol{z}^w\) and preserving generation diversity.

\subsection{Detection and Traceability}

Detection relies primarily on the first stage due to the risk of error propagation in two-stage decoding. Leveraging the larger norms of projections afforded by TTS, the test statistic is computed based on the $L_1$ norm of the first-stage projection vector:
\begin{align}
\label{eq:detection_statistic_pk}
l = \bigl\|\boldsymbol{p}_k\bigr\|_1, \quad (\boldsymbol{p}_k)_j = \langle\widehat{\boldsymbol{z}}^k,\boldsymbol{v}_{kj}\rangle, \quad j=1,\dots,m_k.
\end{align}
In this equation, $m_k$ denotes the size of the session key. $\boldsymbol{p}_k$ is the projection vector obtained by projecting $\widehat{\boldsymbol{z}}^k$ onto the first stage's normal vectors $\{\boldsymbol{v}_{kj}\}_{j=1}^{m_k}$. This statistic $l$ is compared against a threshold $d$ calibrated to a target false-positive rate.

For traceability, full two-stage decoding is performed to recover the complete embedded bit-vector \(\widehat{\boldsymbol{b}}\). This decoded watermark is compared against the registered identity watermark database. The account whose assigned watermark yields the highest similarity score (e.g., match count or Hamming distance) is identified as the image's originator, enabling misuse tracing and accountability.

\section{Experiments}
\label{sec:exp}

\subsection{Experimental Setup}
\label{sec:setting}

\paragraph{Implementation Details.}  
Our image generation backbone is Stable Diffusion v2.1 (SD v2.1) \cite{ldm_2022}, configured with a guidance scale of 7.5, 50 DDIM denoising steps, and a fixed \(512\times512\) output resolution. T2SMark employs a 16-bit session key and a 256-bit watermark. We set the truncation threshold to \( \tau=0.674\). 
The session key is embedded in the first channel of the initial noise. 
This is determined empirically in our parameter selection study (see Appendix \ref{sec:tau}). 
All the experiments are implemented in PyTorch 2.4.1 and run on a single NVIDIA RTX A6000 GPU.

\paragraph{Baselines.}  
We compare against three categories of existing methods. Traditional post-processing transforms include dwtDct \cite{dwt2007}, dwtDctSvd \cite{dwt2007}, and the learning-based RivaGAN \cite{rivagan2019}; fine-tuning approaches are represented by Stable Signature \cite{stablesig_2023}; Inversion-based Schemes include Tree–Ring (TRW) \cite{treering_2023}, Gaussian Shading (GS) \cite{gs_2024} and PRC-Watermark (PRCW) \cite{prcw_2025}. For all inversion-based methods, we perform 10‐step DDIM inversion \cite{ddim_2021}. During inversion, we employ an empty prompt and fix the guidance scale at 1 to simulate unknown prompt conditions. To ensure fair capacity, dwtDct, dwtDctSvd, Gaussian Shading, and PRC-Watermark all embed 256 bits. RivaGAN and Stable Signature use 32 bits and 48 bits respectively, following their official implementations.

\begin{figure}[htbp]
    \centering
    \includegraphics[width=0.75\linewidth]{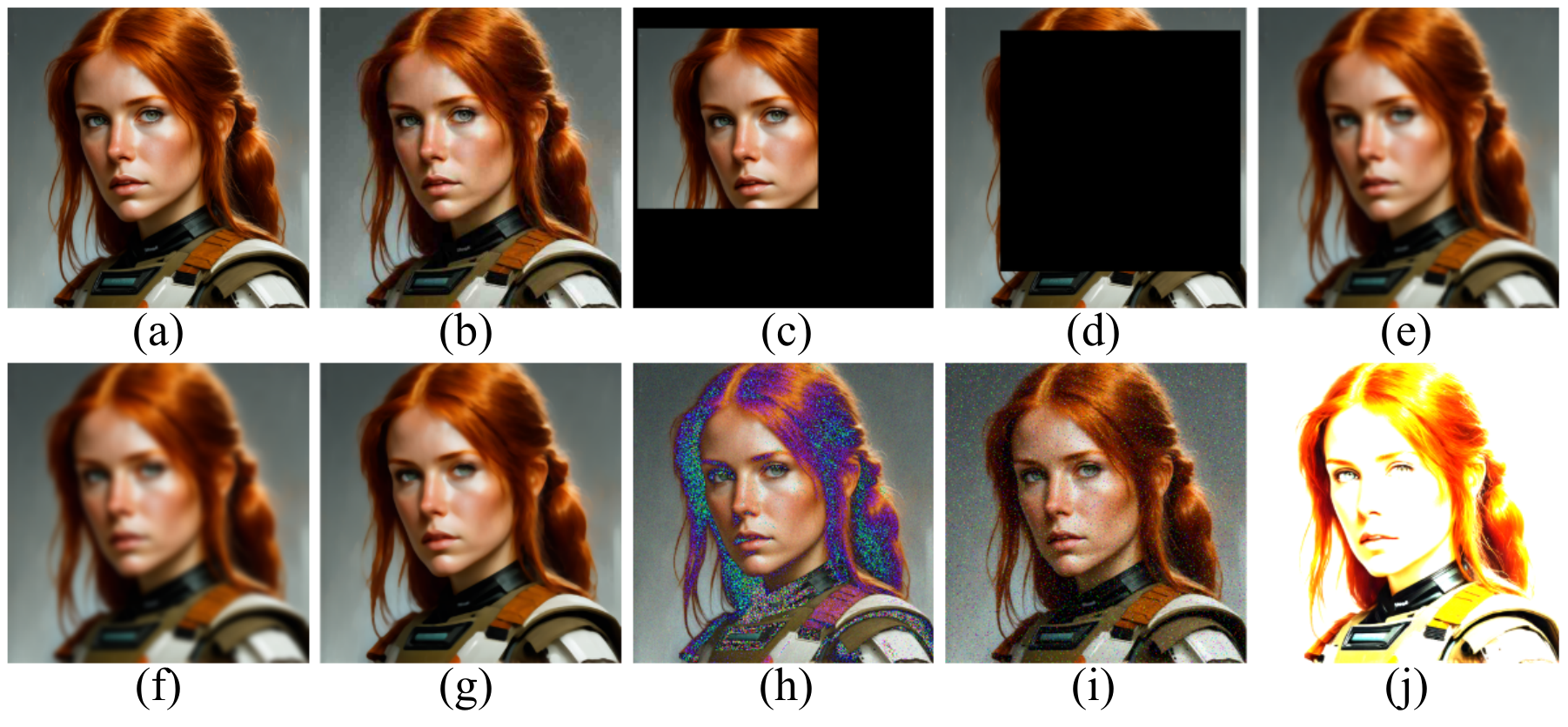}
\caption{The watermarked image is attacked by different types of noise. (a) Watermarked image. (b) JPEG, \(QF\) = 25. (c) 60\% area Random Crop (RandCr). (d) 80\% area Random Drop (RandDr). (e) 25\% Resize and restore (Resize). (f) Gaussian Blur, \(r\) = 4 (GauBlur). (g) Median Blur, \(k\) = 7 (MedBlur). (h) Gaussian Noise, \(\mu\) = 0, \(\sigma\) = 0.05 (GauNoise). (i) Salt and Pepper Noise, \(p\) = 0.05 (S\&PNoise).  (j) Brightness, \(factor\) = 6.}
    \label{fig:noise}
\end{figure}

\paragraph{Evaluation.}  
We evaluate on MS-COCO-2017 \cite{coco_2014} dataset (COCO) and Stable-Diffusion-Prompt\footnote{\href{https://huggingface.co/datasets/Gustavosta/Stable-Diffusion-Prompts}{Stable-Diffusion-Prompts}} dataset (SDP). 
For robustness, we compare the TPR at a fixed \(\text{FPR}=10^{-6}\) in the detection setting and per-bit accuracy in the traceability setting. For each method, we sample 500 prompts from the SDP training split, generate 500 watermarked images, apply nine different distortions (see Figure \ref{fig:noise}), and then perform detection and traceability.

We assess generation diversity via LPIPS \cite{lpips_2018}. For each non–post-processing method, we generate 10 images for each of 1{,}000 COCO test prompts with a fixed 
master key and watermark, compute LPIPS over the 45 unique pairs per prompt, and report the overall mean. Post-processing methods leave the core generation unchanged and are therefore omitted from this evaluation.

For visual quality, we report the CLIP score \cite{clip_2021} and FID \cite{fid_2017}. We run 10 independent trials: in each trial, we fix a single key–watermark pair (simulating one user), and generate 1\,000 images from COCO’s test prompts for the CLIP score and FID. We then perform two-sample \(t\)-tests to assess the impact of the watermark on quality. The hypotheses are 
\(
H_0: \mu_s = \mu_0,
\quad
H_1: \mu_s \neq \mu_0,
\)
where \(\mu_s\) and \(\mu_0\) denote the mean FID or CLIP score computed over multiple sets of watermarked and clean images, respectively. A lower \(t\)-value indicates stronger support for \(H_0\). 
Details about the \(t\)-test can be found in Appendix \ref{sec:ttest_about}.

\subsection{Main Results}

\begin{table}[htbp]
\centering
\caption{Evaluation results of watermarking methods, including detection (TPR), traceability (Bit Acc.), generation diversity, and visual quality (CLIP score and FID). 
TPR and Bit Acc. are shown as Clean/Adv.
Standard errors and $t$-values are shown for the CLIP score and FID.}
\label{tab:results21}
\resizebox{\textwidth}{!}{%
\begin{tabular}{lcccccc}
\toprule
Method & TPR & Bit Acc. & Diversity \(\uparrow\) & CLIP Score (\(t\downarrow\)) & FID (\(t\downarrow\)) \\
\midrule
SD v2.1 \cite{ldm_2022}      
  & –                  & –                       & 0.7072     & 0.3224\(\pm\)0.0010 (--)     & 56.8132\(\pm\)0.48 (--)      \\
\midrule
dwtDct \cite{dwt2007}        
  & 0.922/0.173        & 0.8177/0.5744           & –          & 0.3206\(\pm\)0.0010 (3.7233)& 56.3471\(\pm\)0.48 (2.0270)  \\
dwtDctSvd \cite{dwt2007}     
  & \textbf{1.000}/0.471 & 0.9988/0.6800   & –          & 0.3209\(\pm\)0.0010 (3.2150)& 56.0986\(\pm\)0.48 (3.1220)  \\
RivaGAN \cite{rivagan2019}   
  & 0.914/0.436        & 0.9823/0.8666           & –          & 0.3209\(\pm\)0.0010 (3.2150)& 56.1893\(\pm\)0.48 (2.7332)  \\
StableSig \cite{stablesig_2023}
  & \textbf{1.000}/0.418& 0.9981/0.7462           & 0.6917     & 0.3235\(\pm\)0.0007 (2.6063)& 56.0423\(\pm\)0.38 (3.7020)  \\
TRW \cite{treering_2023}      
  & \textbf{1.000}/0.907& –/–                     & 0.6943     & 0.3210\(\pm\)0.0007 (3.6365)& 58.2667\(\pm\)0.36 (6.9293)  \\
GS \cite{gs_2024}            
  & \textbf{1.000}/\textbf{0.998} & \textbf{1.0000}/\underline{0.9548} & 0.6446     & 0.3242\(\pm\)0.0027 (1.7557) & 58.1377\(\pm\)1.19 (3.0807) \\
PRCW \cite{prcw_2025}
  & \textbf{1.000}/0.294& 0.6494/0.5024           & \textbf{0.7074}     & 0.3218\(\pm\)0.0009 (\underline{1.4369}) & 56.8975\(\pm\)0.38 (\textbf{0.4056}) \\
T2SMark                      
  & \textbf{1.000}/\textbf{0.998} & \textbf{1.0000}/\textbf{0.9754} & \underline{0.7069}     & 0.3227\(\pm\)0.0008 (\textbf{0.5081}) & 56.9317\(\pm\)0.42 (\underline{0.5490}) \\
\bottomrule
\end{tabular}%
}
\end{table}

Results for the different methods are presented in Table \ref{tab:results21}. T2SMark demonstrates superior performance in the traceability scenario while achieving detection performance comparable to the best method, GS \cite{gs_2024}. In contrast, PRCW \cite{prcw_2025} performs worst in traceability and registers a detection TPR below 30\% under adversarial conditions, which is too fragile for real-world deployment.
More detailed robustness results under various noise types can be found in Appendix \ref{sec:detailed_robust}.

With respect to generation diversity, measured by LPIPS \cite{lpips_2018}, PRCW achieves the highest score. T2SMark trails closely behind, with a difference of less than \(10^{-3}\), representing a negligible gap. GS exhibits the lowest diversity score, which is attributable to its use of a fixed codeword for each user, while Stable Signature and TRW also experience noticeable reductions in diversity.

Regarding image quality, only T2SMark and PRCW consistently satisfy the no-degradation criterion across both evaluation tests. Although GS achieves a competitive CLIP score \cite{clip_2021}, its FID \cite{fid_2017} deviates significantly from the no-watermark baseline. Furthermore, GS exhibits a noticeably larger standard deviation in the CLIP score and FID, suggesting that its generation quality is sensitive to the specific watermark and key used. This sensitivity poses a challenge for the user experience, potentially leading to inconsistent generation quality when different user accounts (with different watermarks/keys) are involved. Considering all the evaluated metrics, T2SMark achieves the best overall balance.

\subsection{Undetectability}

To evaluate watermark undetectability, we trained a ResNet-18 classifier \cite{resnet_2016} to distinguish between watermarked and non-watermarked images. We assessed four inversion-based methods using a fixed key and watermark for each. The training set for each method consisted of 8,000 watermarked and 8,000 clean samples, while the test set contained 500 samples each. Training was conducted for 10 epochs with a batch size of 128 and a learning rate of \(1\times10^{-4}\). Table~\ref{tab:undet} reports the test accuracy.
The results indicate that TRW \cite{treering_2023} and GS \cite{gs_2024} are relatively easy to detect. Although PRCW \cite{prcw_2025} demonstrates the highest level of undetectability, we observe that it is not entirely immune to detection. T2SMark achieves the second-best performance and can also be considered difficult to detect, demonstrating a high level of imperceptibility.

\begin{table}[htbp]
\caption{Undetectability of different inversion-based watermarking methods, measured by detection accuracy (Det. Acc.). A lower accuracy indicates better undetectability.}
\label{tab:undet}
\centering
\begin{tabular}{ccccc}
\toprule
 & TRW \cite{treering_2023} & GS \cite{gs_2024} & PRCW \cite{prcw_2025} & T2SMark \\
\midrule
Det. Acc. & 0.971 & 0.994 & \textbf{0.532} & \underline{0.578} \\
\bottomrule
\end{tabular}
\end{table}

\subsection{Generalizability}

We evaluate T2SMark along with other inversion-based watermarking methods on the Stable Diffusion v3.5 Medium model (SD v3.5M) \cite{sd3_2024}, which employs DiT \cite{dit_2023} as its denoising network and features a 16-channel latent space. The detailed settings can be found in Appendix \ref{sec:app_exp_details}.

The results in Table~\ref{tab:results35} show that SD v3.5M delivers superior generative quality so that all methods maintain strong visual fidelity. TRW’s robustness declines markedly compared with its performance on SD v2.1, and GS experiences a similar loss of diversity. PRCW improves traceability but still trails the best method by a wide margin. In contrast, T2SMark achieves the best balance of robustness and diversity, and its undetectability is virtually indistinguishable from that of PRCW.

\begin{table}[htbp]
  \centering
    \caption{Evaluation results of inversion-based watermarking methods on SD v3.5M, including detection (TPR), traceability (Bit Acc.), detection accuracy (Det. Acc.), generation diversity, and visual quality (CLIP score and FID). 
The TPR and Bit Acc. are shown as Clean/Adv.
Standard errors and $t$-values are shown for the CLIP score and FID.}
  \label{tab:results35}
  \resizebox{\linewidth}{!}{
  \begin{tabular}{lcccccc}
    \toprule
    Method
      & TPR
      & Bit Acc.
      & Det. Acc. \(\downarrow\)
      & Diversity \(\uparrow\)
      & CLIP Score (\(t\downarrow\))
      & FID (\(t\downarrow\))
       \\
    \midrule
    SD v3.5M \cite{sd3_2024}
      & – / –
      & – / –
      & -
      & 0.6113
      & 0.3498±.0010 (-)
      & 55.7627±.56 (-)
      \\
    \midrule
   TRW \cite{treering_2023}
      & 0.878 / 0.318
      & – / –
      & 0.984
      & 0.5924
      & 0.3493±.0005 (1.3493)
      & 55.9084±.53 (\underline{0.5699})
      \\
    GS \cite{gs_2024}
      & \textbf{1.000 / 0.990}
      & \underline{0.9994} / \underline{0.9663}
      & 0.991
      &  0.5176
      & 0.3502±.0004 (0.9943)
      & 56.2020±.67 (1.5043)
      \\
    PRCW \cite{prcw_2025}
      & 0.998 / 0.279
      & 0.9920 / 0.6067
      & \textbf{0.516}
      & \underline{0.6096}
      & 0.3502±.0005 (\underline{0.9726})
      & 55.5673±.49 (0.7902)
      \\
    T2SMark
      & \textbf{1.000} / \underline{0.985}
      & \textbf{1.0000 / 0.9768}
      & \underline{0.518}
      & \textbf{0.6102}
      & 0.3499±.0004 (\textbf{0.0991})
      & 55.8121±.49 (\textbf{0.1983})
      \\
    \bottomrule
  \end{tabular}}
\end{table}

\subsection{Ablation Study}

In this section we conduct ablation experiments on SD v2.1 to evaluate robustness under varying hyperparameter settings. Unless otherwise specified, we generate 500 images on the SDP dataset and report the true positive rate at a fixed false positive rate of \(10^{-6}\) alongside bit accuracy.

\paragraph{Tail-Truncated Sampling.}
We evaluate T2SMark both with and without Tail-Truncated Sampling (TTS), and additionally measure generation diversity over 1,000 prompts from the SDP dataset. The results in Table~\ref{tab:tts_results} demonstrate that TTS provides a substantial robustness boost while having only a negligible effect on diversity. This confirms TTS as a critical component of T2SMark.

\begin{table}[htbp]
  \centering
  \caption{Performance of T2SMark both with and without Tail-Truncated Sampling.}
  \label{tab:tts_results}
  \begin{tabular}{lccc}
    \toprule
              & TPR (Clean/Adv.)    & Bit Acc. (Clean/Adv.)    &  Diversity \(\uparrow\)          \\
    \midrule
    w/o TTS       & 1.000/0.996 & 0.9988/0.9307 & 0.6743 \\
    w/ TTS        & 1.000/0.998 & 1.0000/0.9754 & 0.6746 \\
    \bottomrule
  \end{tabular}
\end{table}

\begin{figure}[h]
    \centering
    \includegraphics[width=1\linewidth]{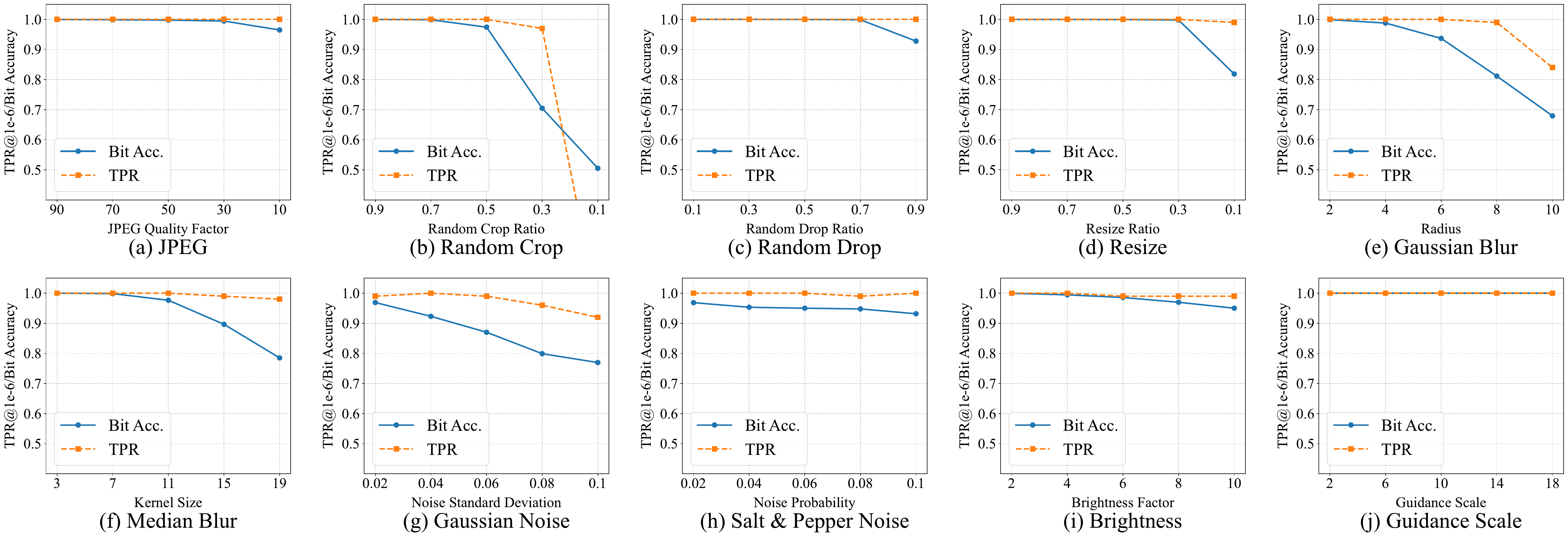}
    \caption{Ablation Studies.}
    \label{fig:robustness_analysis}
\end{figure}

\paragraph{Noise Intensities.}
To further test the robustness, we conduct experiments using different intensities of noise. The results are shown in 
Figure~\ref{fig:robustness_analysis}(a-i).
T2SMark is highly vulnerable to Gaussian noise—even at a low noise standard deviation of 0.1, its performance degrades dramatically, a weakness that has also been reported in experiments of other inversion-based methods \cite{gs_2024, treering_2023}.

\paragraph{Guidance Scales.}

Given diverse user preferences for prompt adherence, higher guidance scales enforce the original prompt more strictly, while lower scales allow greater creative freedom. Since Stable Diffusion v2.1 typically uses guidance scales between 5 and 15, our experiments span a wider range of 2-18. As shown in Figure~\ref{fig:robustness_analysis}(j), T2SMark’s performance degrades only marginally under these settings.

\paragraph{Inversion Steps.}
We evaluate T2SMark with 5, 10, 25, 50, and 100 inversion steps (Table~\ref{tab:inv-steps}) and observe only minor performance fluctuations. Therefore, we can directly improves the extraction efficiency by reducing the number of inversion steps, as inversion is the primary bottleneck in NaW extraction pipelines.

\begin{table}[h]
  \centering
  \begin{minipage}[b]{0.52\textwidth}
    \centering
    \caption{Impact of inversion steps on robustness performance. The TPR and Bit Acc. are shown as Clean/Adv.}
    \label{tab:inv-steps}
    \begin{tabular}{ccc}
      \toprule
      Inversion Steps & TPR & Bit Acc.  \\
      \midrule
      5    & 1.000/0.996 & 1.0000/0.9708 \\
      10   & 1.000/0.997 & 1.0000/0.9764 \\
      25   & 1.000/0.996 & 1.0000/0.9764 \\
      50   & 1.000/0.998 & 1.0000/0.9745 \\
      100  & 1.000/0.997 & 1.0000/0.9730 \\
      \bottomrule
    \end{tabular}
    
  \end{minipage}
  \hfill
  \begin{minipage}[b]{0.4\textwidth}
    \centering
    \caption{Performance under different capacity settings. The Bit Acc. is shown as Clean/Adv.}
    \label{tab:msg_results}
    \begin{tabular}{cc}
      \toprule
      Capacity (bits) & Bit Acc.  \\
      \midrule
      256   & 1.0000/0.9754 \\
      384   & 1.0000/0.9595 \\
      512   & 1.0000/0.9437 \\
      768   & 0.9992/0.9145 \\
      1024  & 0.9968/0.8789 \\
      \bottomrule
    \end{tabular}
    
  \end{minipage}
\end{table}

\paragraph{Watermark Capacity.}

We evaluate T2SMark using watermark capacity of 256, 384, 512, 768 and 1024 bits (Table \ref{tab:msg_results}).
As expected, the robustness of T2SMark degrades with increased capacity. However, if the actual noise level in deployment is lower than that used in our experiments, a moderate increase in capacity remains tolerable.

\paragraph{Session Key Size.}

\begin{table}[htbp]
  \centering
  \caption{Performance under different session key sizes.}
  \label{tab:seed_results}
  \begin{tabular}{lcccc}
    \toprule
              & 8             & 16            & 24            & 32            \\
    \midrule
    TPR (Clean/Adv.)             & 1.000/0.998   & 1.000/0.998   & 1.000/0.996   & 1.000/0.991   \\
    Bit Acc. (Clean/Adv.)        & 1.0000/0.9776 & 1.0000/0.9754 & 1.0000/0.9687 & 1.0000/0.9481 \\
    \bottomrule
  \end{tabular}
\end{table}

We evaluate T2SMark via random key sizes of 8, 16, 24, and 32 bits (Table \ref{tab:seed_results}).
As the length of the random key increases, it becomes more difficult to extract the key perfectly in the first stage, which leads to cascading errors and a rapid drop in bit accuracy. While longer keys can theoretically enhance randomness, we find that a 16-bit key provides sufficient entropy for practical use—since it is unlikely that a single user would generate a number of images large enough to exhaust this space. Moreover, even with a 32-bit random key, T2SMark still maintains acceptable robustness.

\section{Limitations}
\label{sec:limitations}
Despite its wonderful performance on diversity and robustness, T2SMark is subject to several limitations.
A key set of challenges is shared among most NaW methods. First, the high robustness that allows for watermark recovery can be exploited in a forgery attack. An adversary could use a proxy model to invert the diffusion process, recover the watermarked noise vector, and then use this vector to generate new, forged images that appear authentic because they carry a valid watermark \cite{forgery2024}. Second, T2SMark, like other NaW schemes, depends on an invertible, ODE-based sampling method (e.g., DDIM \cite{ddim_2021}, DPM-Solver \cite{dpm_2022}). Without such a sampler, the initial noise cannot be reconstructed, preventing watermark recovery and limiting applicability to diffusion models that support this feature. Third, NaW methods are vulnerable to geometric distortions since their designs lack explicit mechanisms to resist it. Recent work like GaussMarker \cite{gaussmarker_2025} offers effective solutions to this. Finally, there is a potential conflict with certain controllable generation methods that require modifying the core sampling logic, as this could interfere with the watermark embedding and detection process.
In addition to these shared challenges, T2SMark has a unique vulnerability stemming from its key embedding strategy: Embedding the session key in the truncated tails concentrates energy in a way that introduces subtle distributional anomalies, which can be detected.

\section{Conclusion}
T2SMark presents a novel watermarking framework for diffusion-generated images. It leverages Tail-Truncated Sampling (TTS) to increase robustness and employs a two-stage key hierarchy to introduce controlled randomness and support generation diversity. Experiments on both the U-Net \cite{unet_2015} and DiT \cite{dit_2023} architectures demonstrate that T2SMark achieves superior traceability and competitive detection performance while effectively preserving diversity and image quality.

\section*{Acknowledgments}
This work was supported in part by the National Natural Science Foundation of China under Grant 62472398, Grant U2336206, Grant U2436601, and Grant 62402469.

\bibliography{neurips_2025}

\begin{thebibliography}{36}
\providecommand{\natexlab}[1]{#1}
\providecommand{\url}[1]{\texttt{#1}}
\expandafter\ifx\csname urlstyle\endcsname\relax
  \providecommand{\doi}[1]{doi: #1}\else
  \providecommand{\doi}{doi: \begingroup \urlstyle{rm}\Url}\fi

\bibitem[Ho et~al.(2020)Ho, Jain, and Abbeel]{ddpm_2020}
Jonathan Ho, Ajay Jain, and Pieter Abbeel.
\newblock Denoising diffusion probabilistic models.
\newblock In \emph{Advances in Neural Information Processing Systems}, pages 6840--6851. Curran Associates, Inc., 2020.

\bibitem[Song et~al.(2021)Song, Meng, and Ermon]{ddim_2021}
Jiaming Song, Chenlin Meng, and Stefano Ermon.
\newblock Denoising diffusion implicit models.
\newblock In \emph{International Conference on Learning Representations}, 2021.

\bibitem[Rombach et~al.(2022)Rombach, Blattmann, Lorenz, Esser, and Ommer]{ldm_2022}
Robin Rombach, Andreas Blattmann, Dominik Lorenz, Patrick Esser, and Bj\"orn Ommer.
\newblock High-resolution image synthesis with latent diffusion models.
\newblock In \emph{Proceedings of the IEEE/CVF Conference on Computer Vision and Pattern Recognition (CVPR)}, pages 10684--10695, June 2022.

\bibitem[Peebles and Xie(2023)]{dit_2023}
William Peebles and Saining Xie.
\newblock Scalable diffusion models with transformers.
\newblock In \emph{Proceedings of the IEEE/CVF International Conference on Computer Vision (ICCV)}, pages 4195--4205, October 2023.

\bibitem[Yang et~al.(2024)Yang, Zeng, Chen, Fang, Zhang, and Yu]{gs_2024}
Zijin Yang, Kai Zeng, Kejiang Chen, Han Fang, Weiming Zhang, and Nenghai Yu.
\newblock Gaussian shading: Provable performance-lossless image watermarking for diffusion models.
\newblock In \emph{Proceedings of the IEEE/CVF Conference on Computer Vision and Pattern Recognition (CVPR)}, pages 12162--12171, June 2024.

\bibitem[Gunn et~al.(2024)Gunn, Zhao, and Song]{prcw_2025}
Sam Gunn, Xuandong Zhao, and Dawn Song.
\newblock An undetectable watermark for generative image models.
\newblock \emph{arXiv preprint arXiv:2410.07369}, 2024.

\bibitem[Christ and Gunn(2024)]{prc_2024}
Miranda Christ and Sam Gunn.
\newblock Pseudorandom error-correcting codes.
\newblock In \emph{Advances in Cryptology -- CRYPTO 2024}, pages 325--347, 2024.

\bibitem[Esser et~al.(2024)Esser, Kulal, Blattmann, Entezari, M{\"u}ller, Saini, Levi, Lorenz, Sauer, Boesel, et~al.]{sd3_2024}
Patrick Esser, Sumith Kulal, Andreas Blattmann, Rahim Entezari, Jonas M{\"u}ller, Harry Saini, Yam Levi, Dominik Lorenz, Axel Sauer, Frederic Boesel, et~al.
\newblock Scaling rectified flow transformers for high-resolution image synthesis.
\newblock In \emph{Forty-first international conference on machine learning}, 2024.

\bibitem[Ronneberger et~al.(2015)Ronneberger, Fischer, and Brox]{unet_2015}
Olaf Ronneberger, Philipp Fischer, and Thomas Brox.
\newblock U-net: Convolutional networks for biomedical image segmentation.
\newblock In \emph{Medical image computing and computer-assisted intervention--MICCAI 2015: 18th international conference, Munich, Germany, October 5-9, 2015, proceedings, part III 18}, pages 234--241. Springer, 2015.

\bibitem[Wen et~al.(2023)Wen, Kirchenbauer, Geiping, and Goldstein]{treering_2023}
Yuxin Wen, John Kirchenbauer, Jonas Geiping, and Tom Goldstein.
\newblock Tree-ring watermarks: Fingerprints for diffusion images that are invisible and robust.
\newblock \emph{arXiv preprint arXiv:2305.20030}, 2023.

\bibitem[Vaswani et~al.(2017)Vaswani, Shazeer, Parmar, Uszkoreit, Jones, Gomez, Kaiser, and Polosukhin]{transformer2017}
Ashish Vaswani, Noam Shazeer, Niki Parmar, Jakob Uszkoreit, Llion Jones, Aidan~N Gomez, {\L}ukasz Kaiser, and Illia Polosukhin.
\newblock Attention is all you need.
\newblock \emph{Advances in neural information processing systems}, 30, 2017.

\bibitem[Devlin et~al.(2019)Devlin, Chang, Lee, and Toutanova]{bert_2019}
Jacob Devlin, Ming-Wei Chang, Kenton Lee, and Kristina Toutanova.
\newblock Bert: Pre-training of deep bidirectional transformers for language understanding.
\newblock In \emph{Proceedings of the 2019 conference of the North American chapter of the association for computational linguistics: human language technologies, volume 1 (long and short papers)}, pages 4171--4186, 2019.

\bibitem[Brown et~al.(2020)Brown, Mann, Ryder, Subbiah, Kaplan, Dhariwal, Neelakantan, Shyam, Sastry, Askell, et~al.]{gpt_2020}
Tom Brown, Benjamin Mann, Nick Ryder, Melanie Subbiah, Jared~D Kaplan, Prafulla Dhariwal, Arvind Neelakantan, Pranav Shyam, Girish Sastry, Amanda Askell, et~al.
\newblock Language models are few-shot learners.
\newblock \emph{Advances in neural information processing systems}, 33:\penalty0 1877--1901, 2020.

\bibitem[Liu et~al.(2024)Liu, Zhang, Li, Yan, Gao, Chen, Yuan, Huang, Sun, Gao, et~al.]{sora_2024}
Yixin Liu, Kai Zhang, Yuan Li, Zhiling Yan, Chujie Gao, Ruoxi Chen, Zhengqing Yuan, Yue Huang, Hanchi Sun, Jianfeng Gao, et~al.
\newblock Sora: A review on background, technology, limitations, and opportunities of large vision models.
\newblock \emph{arXiv preprint arXiv:2402.17177}, 2024.

\bibitem[Chen et~al.(2023)Chen, Yu, Ge, Yao, Xie, Wu, Wang, Kwok, Luo, Lu, et~al.]{pixart_2023}
Junsong Chen, Jincheng Yu, Chongjian Ge, Lewei Yao, Enze Xie, Yue Wu, Zhongdao Wang, James Kwok, Ping Luo, Huchuan Lu, et~al.
\newblock Pixart-$\backslash{alpha}$: Fast training of diffusion transformer for photorealistic text-to-image synthesis.
\newblock \emph{arXiv preprint arXiv:2310.00426}, 2023.

\bibitem[Lu et~al.(2022{\natexlab{a}})Lu, Zhou, Bao, Chen, LI, and Zhu]{dpm_2022}
Cheng Lu, Yuhao Zhou, Fan Bao, Jianfei Chen, Chongxuan LI, and Jun Zhu.
\newblock Dpm-solver: A fast ode solver for diffusion probabilistic model sampling in around 10 steps.
\newblock In \emph{Advances in Neural Information Processing Systems}, volume~35, pages 5775--5787, 2022{\natexlab{a}}.

\bibitem[Lu et~al.(2022{\natexlab{b}})Lu, Zhou, Bao, Chen, Li, and Zhu]{dpmpp2022}
Cheng Lu, Yuhao Zhou, Fan Bao, Jianfei Chen, Chongxuan Li, and Jun Zhu.
\newblock Dpm-solver++: Fast solver for guided sampling of diffusion probabilistic models.
\newblock \emph{arXiv preprint arXiv:2211.01095}, 2022{\natexlab{b}}.

\bibitem[Cox et~al.(2007)Cox, Miller, Bloom, Fridrich, and Kalker]{dwt2007}
Ingemar Cox, Matthew Miller, Jeffrey Bloom, Jessica Fridrich, and Ton Kalker.
\newblock \emph{Digital watermarking and steganography}.
\newblock Morgan kaufmann, 2007.

\bibitem[Zhang et~al.(2019)Zhang, Xu, Cuesta-Infante, and Veeramachaneni]{rivagan2019}
Kevin~Alex Zhang, Lei Xu, Alfredo Cuesta-Infante, and Kalyan Veeramachaneni.
\newblock Robust invisible video watermarking with attention.
\newblock \emph{arXiv preprint arXiv:1909.01285}, 2019.

\bibitem[Jia et~al.(2021)Jia, Fang, and Zhang]{mbrs_2021}
Zhaoyang Jia, Han Fang, and Weiming Zhang.
\newblock Mbrs: Enhancing robustness of dnn-based watermarking by mini-batch of real and simulated jpeg compression.
\newblock In \emph{Proceedings of the 29th ACM International Conference on Multimedia}, page 41–49, 2021.

\bibitem[Ma et~al.(2022)Ma, Guo, Hou, Yang, Li, Jia, and Xie]{cin_2022}
Rui Ma, Mengxi Guo, Yi~Hou, Fan Yang, Yuan Li, Huizhu Jia, and Xiaodong Xie.
\newblock Towards blind watermarking: Combining invertible and non-invertible mechanisms.
\newblock In \emph{MM '22}, page 1532–1542, 2022.

\bibitem[Fernandez et~al.(2023)Fernandez, Couairon, J\'egou, Douze, and Furon]{stablesig_2023}
Pierre Fernandez, Guillaume Couairon, Herv\'e J\'egou, Matthijs Douze, and Teddy Furon.
\newblock The stable signature: Rooting watermarks in latent diffusion models.
\newblock In \emph{Proceedings of the IEEE/CVF International Conference on Computer Vision (ICCV)}, pages 22466--22477, October 2023.

\bibitem[Min et~al.(2024)Min, Li, Chen, and Cheng]{wadiff_2024}
Rui Min, Sen Li, Hongyang Chen, and Minhao Cheng.
\newblock A watermark-conditioned diffusion model for ip protection.
\newblock In \emph{European Conference on Computer Vision}, pages 104--120. Springer, 2024.

\bibitem[Kingma et~al.(2013)Kingma, Welling, et~al.]{vae_2023}
Diederik~P Kingma, Max Welling, et~al.
\newblock Auto-encoding variational bayes.
\newblock \emph{arXiv preprint arXiv:1312.6114}, 2013.

\bibitem[Ci et~al.(2024)Ci, Yang, Song, and Shou]{ringid_2024}
Hai Ci, Pei Yang, Yiren Song, and Mike~Zheng Shou.
\newblock Ringid: Rethinking tree-ring watermarking for enhanced multi-key identification.
\newblock In \emph{Proceedings of the European Conference on Computer Vision (ECCV)}, page 338–354, 2024.

\bibitem[Huang et~al.(2024)Huang, Wu, and Wang]{robin_2024}
Huayang Huang, Yu~Wu, and Qian Wang.
\newblock Robin: Robust and invisible watermarks for diffusion models with adversarial optimization.
\newblock \emph{Advances in Neural Information Processing Systems}, 37:\penalty0 3937--3963, 2024.

\bibitem[Zhang et~al.(2024)Zhang, Liu, Martin, Bearfield, Brun, and Guan]{zodiac_2024}
Lijun Zhang, Xiao Liu, Antoni~V Martin, Cindy~X Bearfield, Yuriy Brun, and Hui Guan.
\newblock Attack-resilient image watermarking using stable diffusion.
\newblock \emph{Advances in Neural Information Processing Systems}, 37:\penalty0 38480--38507, 2024.

\bibitem[Lin et~al.(2014)Lin, Maire, Belongie, Hays, Perona, Ramanan, Doll{\'a}r, and Zitnick]{coco_2014}
Tsung-Yi Lin, Michael Maire, Serge Belongie, James Hays, Pietro Perona, Deva Ramanan, Piotr Doll{\'a}r, and C~Lawrence Zitnick.
\newblock Microsoft coco: Common objects in context.
\newblock In \emph{Computer vision--ECCV 2014: 13th European conference, zurich, Switzerland, September 6-12, 2014, proceedings, part v 13}, pages 740--755. Springer, 2014.

\bibitem[Zhang et~al.(2018)Zhang, Isola, Efros, Shechtman, and Wang]{lpips_2018}
Richard Zhang, Phillip Isola, Alexei~A Efros, Eli Shechtman, and Oliver Wang.
\newblock The unreasonable effectiveness of deep features as a perceptual metric.
\newblock In \emph{Proceedings of the IEEE conference on computer vision and pattern recognition}, pages 586--595, 2018.

\bibitem[Radford et~al.(2021)Radford, Kim, Hallacy, Ramesh, Goh, Agarwal, Sastry, Askell, Mishkin, Clark, et~al.]{clip_2021}
Alec Radford, Jong~Wook Kim, Chris Hallacy, Aditya Ramesh, Gabriel Goh, Sandhini Agarwal, Girish Sastry, Amanda Askell, Pamela Mishkin, Jack Clark, et~al.
\newblock Learning transferable visual models from natural language supervision.
\newblock In \emph{International conference on machine learning}, pages 8748--8763. PmLR, 2021.

\bibitem[Heusel et~al.(2017)Heusel, Ramsauer, Unterthiner, Nessler, and Hochreiter]{fid_2017}
Martin Heusel, Hubert Ramsauer, Thomas Unterthiner, Bernhard Nessler, and Sepp Hochreiter.
\newblock Gans trained by a two time-scale update rule converge to a local nash equilibrium.
\newblock \emph{Advances in neural information processing systems}, 30, 2017.

\bibitem[He et~al.(2016)He, Zhang, Ren, and Sun]{resnet_2016}
Kaiming He, Xiangyu Zhang, Shaoqing Ren, and Jian Sun.
\newblock Deep residual learning for image recognition.
\newblock In \emph{Proceedings of the IEEE Conference on Computer Vision and Pattern Recognition (CVPR)}, June 2016.

\bibitem[M{\"u}ller et~al.(2024)M{\"u}ller, Lukovnikov, Thietke, Fischer, and Quiring]{forgery2024}
Andreas M{\"u}ller, Denis Lukovnikov, Jonas Thietke, Asja Fischer, and Erwin Quiring.
\newblock Black-box forgery attacks on semantic watermarks for diffusion models.
\newblock \emph{arXiv preprint arXiv:2412.03283}, 2024.

\bibitem[Li et~al.(2025)Li, Huang, Hou, and Hong]{gaussmarker_2025}
Kecen Li, Zhicong Huang, Xinwen Hou, and Cheng Hong.
\newblock Gaussmarker: Robust dual-domain watermark for diffusion models.
\newblock \emph{arXiv preprint arXiv:2506.11444}, 2025.

\bibitem[Deng et~al.(2009)Deng, Dong, Socher, Li, Li, and Fei-Fei]{imagenet2009}
Jia Deng, Wei Dong, Richard Socher, Li-Jia Li, Kai Li, and Li~Fei-Fei.
\newblock Imagenet: A large-scale hierarchical image database.
\newblock In \emph{2009 IEEE conference on computer vision and pattern recognition}, pages 248--255. Ieee, 2009.

\bibitem[Liu et~al.(2022)Liu, Gong, and Liu]{flow_2022}
Xingchao Liu, Chengyue Gong, and Qiang Liu.
\newblock Flow straight and fast: Learning to generate and transfer data with rectified flow.
\newblock \emph{arXiv preprint arXiv:2209.03003}, 2022.

\end{thebibliography}

\newpage
\appendix

\section{BEP Analysis under Tail‐Truncated Sampling}
\label{sec:ber}

We model decoding as a binary decision under additive white Gaussian noise (AWGN) with variance \(\sigma^2\). Owing to the symmetry of the problem, we analyze the case corresponding to the transmission of bit 1 without loss of generality. Let \(\mu(\tau)\) and \(D(\tau)\) represent the mean and variance of a standard normal variable truncated in \([\tau,+\infty)\). Aggregating \(r\) truncated samples yields a test statistic with an effective signal-to-noise ratio
\begin{align}
\mathcal{S}(\tau)
&= \frac{\sqrt{r}\,\mu(\tau)}{\sqrt{D(\tau) + \sigma^2}},
\quad
r = \lfloor 2\,\Phi(-\tau)\,n/m \rfloor \approx 2\,\Phi(-\tau)\,n/m.
\label{eq:A_def}
\end{align}
In practice we approximate \(\mathcal{S}(\tau)\) by  
\begin{align}
A(\tau)
= a\,\frac{\sqrt{\Phi(-\tau)}\,\mu(\tau)}{\sqrt{D(\tau) + \sigma^2}},\quad a = \sqrt{2n/m}
\end{align}
and the corresponding bit‐error probability is  
\begin{align}
P_e(\tau)
= \Phi\bigl(-A(\tau)\bigr),
\label{eq:Pe_def}
\end{align}
where \(\Phi(\cdot)\) is the standard normal CDF.  For a tail‐truncated Gaussian distribution,
\begin{align}
\mu(\tau)
= \mathbb{E} \left[ Z\,|\,Z>\tau \right]
= \frac{\phi(-\tau)}{\Phi(-\tau)},
\quad
D(\tau)
= \operatorname{Var} \left[ Z\,|\,Z>\tau \right]
= 1 + \tau\,\mu(\tau) - \mu(\tau)^2,
\end{align}
where \(\phi(\cdot)\) is the standard normal PDF.

To assess whether the truncation’s SNR gain outweighs the redundancy loss for small \(\tau>0\), define  
\begin{align}
B(\tau)
= \frac{\phi(-\tau)}{\sqrt{\Phi(-\tau)}},
\qquad
C(\tau)
= \frac{1}{\sqrt{D(\tau)+\sigma^2}}.
\label{eq:BC_def}
\end{align}
Then \(A(\tau)=a\,B(\tau)\,C(\tau)\) and  
\[
P_e'(\tau)
= -\phi\bigl(-A(\tau)\bigr)\,A'(\tau),
\quad
A'(\tau)
= a\bigl[B'(\tau)\,C(\tau) + B(\tau)\,C'(\tau)\bigr].
\]
First, at \(\tau=0\),
\begin{align}
B(0)
&= \frac{\phi(0)}{\sqrt{\Phi(0)}}
= \frac{1/\sqrt{2\pi}}{\sqrt{1/2}}
= \frac{1}{\sqrt{\pi}}>0,
\label{eq:B0_sign_wrap}\\
C(0)
&= \frac{1}{\sqrt{D(0)+\sigma^2}}>0,
\label{eq:C0_sign_wrap}
\end{align}
since \(D(0)\ge0\) and \(\sigma^2>0\).  Next,
\begin{align}
B'(0)
&= \frac{d}{d\tau}\bigl[\phi(-\tau)\,\Phi(-\tau)^{-1/2}\bigr]_{\tau=0}
= \frac12\,\frac{\phi(0)^2}{\Phi(0)^{3/2}}
= \frac{\sqrt{2}}{2\pi}>0,
\label{eq:Bprime0_sign_wrap}\\
\mu'(0)
&= \frac{d}{d\tau}\bigl[\phi(-\tau)\,\Phi(-\tau)^{-1}\bigr]_{\tau=0}
= \frac{\phi(0)^2}{\Phi(0)^2}
= \frac{2}{\pi}>0,
\label{eq:muprime0_sign_wrap}\\
D'(0)
&= \bigl[\mu(\tau)+\tau\mu'(\tau)-2\,\mu(\tau)\,\mu'(\tau)\bigr]_{\tau=0}
= \sqrt{\tfrac{2}{\pi}}\Bigl(1-\tfrac{4}{\pi}\Bigr)
<0.
\label{eq:Dprime0_sign_wrap}
\end{align}
Since \(C'(\tau)=-\tfrac12\,(D(\tau)+\sigma^2)^{-3/2}\,D'(\tau)\), and from \(\eqref{eq:C0_sign_wrap}\) and \(\eqref{eq:Dprime0_sign_wrap}\) we have
\begin{align}
C'(0)
= -\tfrac12\underbrace{(D(0)+\sigma^2)^{-3/2}}_{>0}\,
      \underbrace{D'(0)}_{<0}
>0.
\label{eq:Cprime0_sign_wrap}
\end{align}
Combining \(\eqref{eq:B0_sign_wrap}\), \(\eqref{eq:C0_sign_wrap}\), \(\eqref{eq:Bprime0_sign_wrap}\), and \(\eqref{eq:Cprime0_sign_wrap}\) in the product rule for \(A'(\tau)\) gives
\begin{align}
A'(0)= a\Bigl[\underbrace{B'(0)\,C(0)}_{>0}
      + \underbrace{B(0)\,C'(0)}_{>0}\Bigr]
>0,
\quad
P_e'(0)=-\underbrace{\phi\bigl(-A(0)\bigr)}_{>0}\,\underbrace{A'(0)}_{>0}<0.
\end{align}
This analysis, following the provided equations, demonstrates that under the AWGN model, initiating truncation (i.e.\ using a small threshold \(\tau>0\)) strictly decreases the bit‐error probability.  Although this model is an idealization, it predicts the existence of an optimal threshold \(\tau\), whose value is typically determined empirically, as discussed in Section~\ref{sec:tau}.

\section{Experimental Details and Additional Experiments}

\subsection{Implementation Details}
\label{sec:app_exp_details}

\paragraph{Experiments on SD v2.1.}
In these experiments, we use Stable Diffusion v2.1 \cite{ldm_2022} to generate \(512\times512\) images with a guidance scale of 7.5 and 50 DDIM denoising steps \cite{ddim_2021}, into which we embed watermarks. For dwtDct, dwtDctSvd \cite{dwt2007}, and RivaGAN \cite{rivagan2019}, we employ the implementation from the GitHub repository\footnote{\url{https://github.com/ShieldMnt/invisible-watermark}}. Stable Signature \cite{stablesig_2023}, TRW \cite{treering_2023}, GS \cite{gs_2024}, and PRCW \cite{prcw_2025} are run via their official GitHub codes. We fine‐tune Stable Signature for 100 steps with a batch size of 4 on 400 images from the ImageNet2014 validation set \cite{imagenet2009}. Since TRW is a single‐bit scheme, we evaluate it only in the detection setting, using its robust \textit{Ring} mode in channel 3 as recommended. GS is used with its default paper settings, and PRCW is configured for 256 bit capacity. RivaGAN and Stable Signature retain their original capacities of 32 and 48 bits, respectively—higher capacities generally worsen extraction performance and image quality. However, none of these methods surpass T2SMark even under these generous capacity settings.

\paragraph{Experiments on SD v3.5M.} 

To assess the generalizability of our proposed T2SMark, we evaluate it alongside other inversion‐based watermarking methods on the Stable Diffusion 3.5 Medium model (SD v3.5M) \cite{sd3_2024}. SD v3.5 adopts a Rectified Flow training objective \cite{flow_2022}, operates in a larger 16-channel latent space, and leverages the more scalable DiT transformer architecture. Because it also employs ODE-based sampling, inversion-based watermarking methods can be adapted to it directly. For robustness, we generate watermarked images for 500 prompts from the SDP training set with a guidance scale of 7.5; for image quality, we sample the COCO test set \cite{coco_2014} with a guidance scale of 4.0 to emphasize diversity. All methods use 40 inference steps and 10 inversion steps. Embedding occurs in the 16-channel latent space: for TRW and T2SMark’s first stage we use the first four channels; GS uses \(f_{ch}=4\), \(f_h=8\), and \(f_w=8\) (256-bit capacity); and PRCW and T2SMark both embed a 256-bit watermark. T2SMark further employs a 16-bit random key, as in our SD v2.1 setup.

\subsection{Parameter Selection}

\paragraph{Threshold \(\tau\).}
\label{sec:tau}
We select \(\tau\) by generating 256-bit, single-stage watermarks on 100 non-overlapping SDP prompts, applying distortions, and measuring bit accuracy. As shown in Table \ref{tab:tau_accuracy}, bit accuracy first increases rapidly with \(\tau\) and then slowly decreases. To balance robustness without over-constraining sampling, we choose
\(\tau = -\Phi^{-1}\Bigl(\tfrac{4}{16}\Bigr) \approx 0.674.\)

\begin{table}[h]
  \centering
  \caption{Bit accuracy under different truncation thresholds \(\tau\).}
  \label{tab:tau_accuracy}
  \resizebox{\linewidth}{!}{
  \begin{tabular}{ccccccccc}
    \toprule
    \(\tau\)  
      & 0 & \(-\Phi^{-1}\bigl(\tfrac{7}{16}\bigr)\) 
      & \(-\Phi^{-1}\bigl(\tfrac{6}{16}\bigr)\) 
      & \(-\Phi^{-1}\bigl(\tfrac{5}{16}\bigr)\) 
      & \(-\Phi^{-1}\bigl(\tfrac{4}{16}\bigr)\) 
      & \(-\Phi^{-1}\bigl(\tfrac{3}{16}\bigr)\) 
      & \(-\Phi^{-1}\bigl(\tfrac{2}{16}\bigr)\) 
      & \(-\Phi^{-1}\bigl(\tfrac{1}{16}\bigr)\) \\
    \midrule
    Bit Acc.
      & 0.9447 & 0.9828 & 0.9873 & 0.9860 & 0.9868 & 0.9855 & 0.9804 & 0.9600 \\
    \bottomrule
  \end{tabular}}
\end{table}

\paragraph{Embedding Channel of the Session Key.}
\label{sec:channel_selection}

\begin{table}[h]
  \centering
    \caption{Post-distortion inversion MSE and detection accuracy (Det. Acc.) by channel.}
  \label{tab:channel21}
  \begin{tabular}{ccccc}
    \toprule
    Channel      & 0      & 1            & 2      & 3      \\
    \midrule
    MSE \(\downarrow\)         & 0.8018 & \textbf{0.7815} & 0.8349 & 0.8352 \\
    Det. Acc. \(\downarrow\) & 0.578  & 0.612 & 0.633  & \textbf{0.553} \\
    \bottomrule
  \end{tabular}

\end{table}

Different latent channels exhibit varying inversion errors and detectability. We measure the post-distortion inversion MSE and detection rate for channels 0–3 on the same dataset. As Table \ref{tab:channel21} shows, Channel 1 achieves the lowest MSE but is easily detected, while Channel 3 has the lowest detection rate but higher error. We therefore select \textbf{Channel 0} to balance undetectability and reconstruction fidelity.

\subsection{Details about the \(t\)-test}
\label{sec:ttest_about}

For evaluating the CLIP score \cite{clip_2021} and FID \cite{fid_2017}, we generated \(n_s = n_0 = 10\) image sets for a two‐sample \(t\)‐test. We test the hypotheses
\begin{align}
H_0: \mu_s = \mu_0,
\quad
H_1: \mu_s \neq \mu_0,
\end{align}
where \(\mu_s\) and \(\mu_0\) denote the mean metric values for watermarked and clean images, respectively. The \(t\)-statistic is computed as
\begin{align}
t = \frac{|\mu_s - \mu_0|}{S^* \sqrt{\frac{1}{n_s} + \frac{1}{n_0}}},
\qquad
S^* = \sqrt{\frac{(n_s - 1)S_s^2 + (n_0 - 1)S_0^2}{n_s + n_0 - 2}},
\end{align}
where \(S_s\) and \(S_0\) are the sample standard deviations. A lower \(t\)‐value supports \(H_0\), while \(t > t_{0.05,\;n_s+n_0-2}=t_{0.05,18}\approx2.101\) leads us to reject \(H_0\) and conclude that watermarking significantly affects model performance. Otherwise, we consider the watermark to be distortion‐free.

\subsection{Detailed Results of Robustness Evaluation}
\label{sec:detailed_robust}

We report robustness evaluation results on SD v2.1 against various noise types in Table~\ref{tab:robust_all21}, with corresponding results on SD v3.5M in Table~\ref{tab:robust_all_35}. T2SMark consistently outperforms all competing methods across nearly every distortion. Its single notable weakness is Gaussian noise, which likely arises because the inversion process is especially sensitive to this perturbation—an effect also observed in other inversion-based schemes \cite{gs_2024, treering_2023}. Cascading errors of the
two-stage framework exacerbate this vulnerability, causing T2SMark to incur a greater performance decline under Gaussian noise.

\begin{table}[htbp]
\caption{Detection (TPR) and traceability (bit accuracy) results for watermarking methods on SD v2.1 under various noise distortions. Values are shown as TPR/Bit Acc.}
\label{tab:robust_all21}
\centering
\resizebox{\textwidth}{!}{
\begin{tabular}{lcccccccc}
\toprule
Noise       & dwtDct \cite{dwt2007}           & dwtDctSvd  \cite{dwt2007}        & RivaGan \cite{rivagan2019}         & StableSig \cite{stablesig_2023}        & TRW \cite{treering_2023}              & GS  \cite{gs_2024}               & PRCW  \cite{prcw_2025}             & T2SMark           \\
\midrule
None        & 0.922/0.8177     & \textbf{1.000}/0.9988 & 0.914/0.9823     & \textbf{1.000}/0.9981 & \textbf{1.000}/–      & \textbf{1.000}/\textbf{1.0000} & \textbf{1.000}/0.6494    & \textbf{1.000}/\textbf{1.0000} \\
JPEG        & 0.000/0.5015     & 0.002/0.5225       & 0.150/0.8091     & 0.384/0.7934       & 0.678/–           & \textbf{1.000}/0.9794    & 0.448/0.5017       & \textbf{1.000}/\textbf{0.9901} \\
RandCr      & 0.986/0.7675     & 0.998/0.7833       & 0.762/0.9476     & 0.994/0.9873       & 0.980/–           & \textbf{1.000}/0.9421    & 0.078/0.4985       & \textbf{1.000}/\textbf{0.9935} \\
RandDr      & 0.010/0.6012     & 0.000/0.6185       & 0.700/0.9352     & 0.984/0.9714       & 0.982/–           & \textbf{1.000}/0.9156    & 0.052/0.5004       & \textbf{1.000}/\textbf{0.9895} \\
Resize      & 0.000/0.5159     & 0.982/0.8301       & 0.740/0.9442     & 0.000/0.5130       & 0.998/–      & \textbf{1.000}/0.9893    & 0.560/0.4983       & \textbf{1.000}/\textbf{0.9963} \\
GauBlur     & 0.000/0.5029     & 0.352/0.6331      & 0.194/0.8171     & 0.000/0.4104       & 0.994/–      & \textbf{1.000}/0.9661    & 0.070/0.4991       & \textbf{1.000}/\textbf{0.9890} \\
MedBlur     & 0.000/0.5251     & 0.998/0.9113       & 0.780/0.9516     & 0.000/0.6511       & 0.982/–      & \textbf{1.000}/0.9945    & 0.748/0.4997       & \textbf{1.000}/\textbf{0.9984} \\
GauNoise    & 0.318/0.6236     & 0.850/0.7807       & 0.250/0.7948     & 0.458/0.7760       & 0.758/–           & 0.986/\textbf{0.9014}   & 0.054/0.4985       & \textbf{0.994}/0.8961    \\
S\&PNoise   & 0.114/0.5878     & 0.000/0.5168       & 0.082/0.7991     & 0.076/0.7020       & 0.948/–           & \textbf{1.000}/0.9347    & 0.072/0.4992       & 0.992/\textbf{0.9456}    \\
Brightness  & 0.132/0.5439     & 0.060/0.5241       & 0.266/0.8009     & 0.866/0.9134       & 0.856/–           & 0.992/0.9704    & 0.560/0.5259       & \textbf{0.994}/\textbf{0.9803} \\
Adv.\ (ave) & 0.173/0.5744     & 0.471/0.6800       & 0.432/0.8666     & 0.418/0.7462       & 0.907/–           & \textbf{0.998}/0.9548    & 0.294/0.5024       & \textbf{0.998}/\textbf{0.9754} \\
\bottomrule
\end{tabular}}
\end{table}

\begin{table}[htbp]
\caption{Detection (TPR) and traceability (bit accuracy) results for watermarking methods on SD v3.5M under various noise distortions. Values are shown as TPR/Bit Acc.}
\label{tab:robust_all_35}
\centering
\begin{tabular}{lcccc}
\toprule
Noise        & TRW\cite{treering_2023}    & GS \cite{gs_2024} & PRCW \cite{prcw_2025} & T2SMark \\
\midrule
None         & 0.878 / -     & \textbf{1.000} / 0.9994    & 0.998 / 0.9920 & \textbf{1.000} / \textbf{0.9998} \\
JPEG         & 0.050 / -     & \textbf{0.998} / 0.9868    & 0.312 / 0.5968 & \textbf{0.998} / \textbf{0.9955} \\
RandCr       & 0.596 / -     & \textbf{0.998} / 0.9894    & 0.482 / 0.6641 & \textbf{0.998} / \textbf{0.9989} \\
RandDr       & 0.822 / -     & 0.992 / 0.9694    & 0.474 / 0.6597 & \textbf{0.998} / \textbf{0.9955} \\
Resize       & 0.030 / -     & \textbf{0.998} / 0.9843    & 0.170 / 0.5596 & \textbf{0.998} /\textbf{ 0.9972} \\
GauBlur      & 0.098 / -     & 0.994 / 0.9487    & 0.004 / 0.5017 & \textbf{0.998} / 0\textbf{.9790} \\
MedBlur      & 0.004 / -     & \textbf{0.998} / 0.9923    & 0.250 / 0.5993 & \textbf{0.998} / \textbf{0.9983} \\
GauNoise     & 0.410 / -     & \textbf{0.946} / \textbf{0.8832}    & 0.038 / 0.5111 & 0.886 / 0.8629 \\
S\&PNoise    & 0.738 / -     & \textbf{0.996} / 0.9528    & 0.026 / 0.5057 & \textbf{0.996} / \textbf{0.9664} \\
Brightness   & 0.116 / -     & 0.994 / 0.9898    & 0.756 / 0.8619 & \textbf{0.998} / \textbf{0.9978} \\
Adv.\ (ave)  & 0.318 / -     & \textbf{0.990} / 0.9663    & 0.279 / 0.6067 & 0.985 / \textbf{0.9768} \\
\bottomrule
\end{tabular}
\end{table}

\subsection{Results on Different Datasets}

To avoid dataset‐induced bias, we evaluate watermarking methods on two distinct datasets. Robustness is measured on the COCO dataset \cite{coco_2014}, while image quality and generation diversity are assessed on the SDP dataset. Since SDP provides no ground‐truth references, the FID cannot be computed; therefore, we report only the CLIP score and LPIPS diversity. The results are summarized in Table~\ref{tab:results21_dataset}.

\begin{table}[htbp]
\centering
\caption{Additional evaluation results of watermarking methods, including detection (TPR), traceability (Bit Acc.), generation diversity, and visual quality (CLIP score). 
TPR and Bit Acc. shown as Clean/Adv.
Standard errors and $t$-values are shown for the CLIP score.
}
\label{tab:results21_dataset}
\resizebox{\textwidth}{!}{%
\begin{tabular}{lccccc}
\toprule
Method & TPR & Bit Acc. & Diversity \(\uparrow\) & CLIP Score (\(t\downarrow\)) \\
\midrule
SD v2.1 \cite{ldm_2022}      
  & –                  & –                       & 0.6756     &  0.3357\(\pm\).0008(--)      \\
\midrule
dwtDct \cite{dwt2007}        
  & 0.938/0.176        & 0.8493/0.5674           & –          &  0.3347\(\pm\).0007(2.9012) \\
dwtDctSvd \cite{dwt2007}     
  & \textbf{1.000}/0.472& 0.9998/0.6739   & –          & 0.3289\(\pm\).0014(12.613)  \\
RivaGAN \cite{rivagan2019}   
  & 0.980/0.511        & 0.9951/0.8829           & –          & 0.3324\(\pm\).0009(8.5932)  \\
StableSig \cite{stablesig_2023}
  & 0.998/0.404& 0.9965/0.7422           & 0.6667     & 0.3327\(\pm\).0009(7.4739)  \\
TRW \cite{treering_2023}      
  & \textbf{1.000}/0.892 & –/–                     & 0.6655     & 0.3393\(\pm\).0007(10.139)  \\
GS \cite{gs_2024}            
  & \textbf{1.000}/\textbf{0.997} & \textbf{1.0000}/\underline{0.9639} & 0.6156     & 0.3354\(\pm\).0035(\textbf{0.2229}) \\
PRCW \cite{prcw_2025}
  & \textbf{1.000}/0.450& 0.7956/0.5068           & \textbf{0.6747}     & 0.3352\(\pm\).0006(1.4729)  \\
T2SMark                      
  & \textbf{1.000}/\underline{0.996} & \textbf{1.0000}/\textbf{0.9789} & \underline{0.6746}     & 0.3351\(\pm\).0012(\underline{0.3634})  \\
\bottomrule
\end{tabular}%
}
\end{table}

\subsection{Details about Evaluation of Undetectability}

\begin{figure}[h]
    \centering
    \includegraphics[width=1\linewidth]{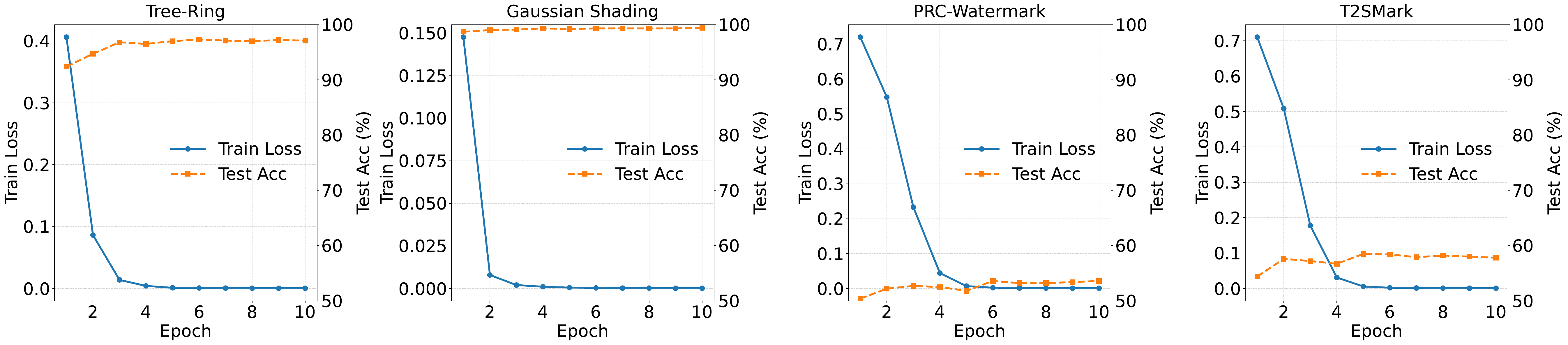}
    \caption{Training Loss and Test Accuracy over Epochs on SD v2.1.}
    \label{fig:train21}
\end{figure}

\begin{figure}[h]
    \centering
    \includegraphics[width=1\linewidth]{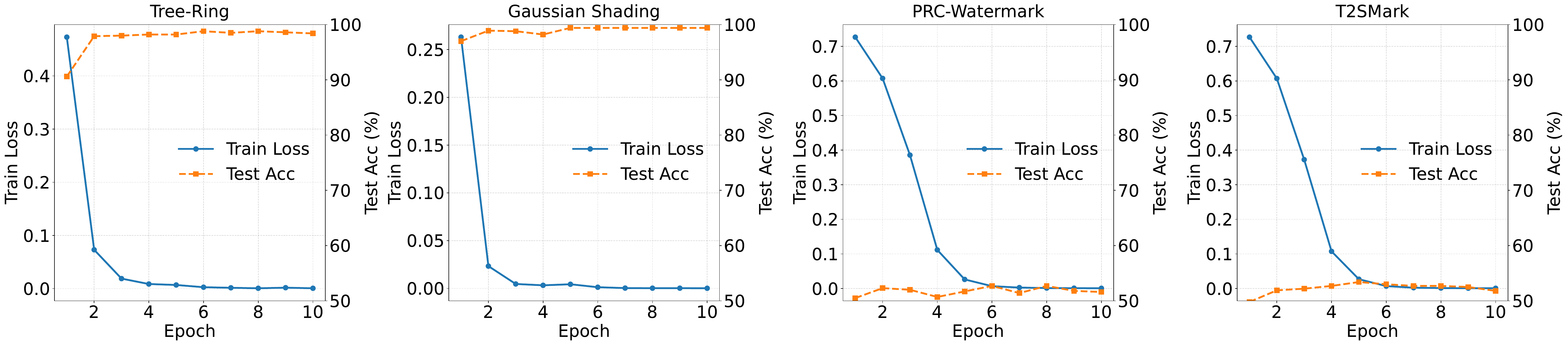}
    \caption{Training Loss and Test Accuracy over Epochs on SD v3.5M.}
    \label{fig:train35}
\end{figure}

To assess undetectability, we fine-tuned a pretrained ResNet-18 \cite{resnet_2016} classifier. For each inversion-based watermarking method and an unwatermarked Stable Diffusion baseline, we generated 8{,}500 images from the SDP training set—8{,}000 for training and 500 for testing. The semantic content was controlled with a fixed prompt set so that the classifier could not exploit content differences. All the watermark methods use a fixed key and watermark across images. Figures~\ref{fig:train21} and~\ref{fig:train35} respectively show the training loss and test accuracy over epochs on SD v2.1 and SD v3.5M. TRW \cite{treering_2023} and GS \cite{gs_2024} are easily detected, while PRCW \cite{prcw_2025} and T2SMark achieve only marginal detection accuracies just above 50\%. Notably, PRCW is not completely immune to detection, with a test accuracy that is consistently greater than 50\%. T2SMark performs better on SD v3.5M, indicating that the diffusion model itself critically impacts undetectability. Purely cryptographic undetectability may over-constrain the encoding and sacrifice robustness.

\subsection{Detailed Results of Ablation Studies}

Tables \ref{tab:detailed_np}, \ref{tab:detailed_cap} and \ref{tab:detailed_key} present detailed ablation results for the inversion steps, watermark capacity and session key size. Within practical ranges, our method supports flexible adjustment of the embedding capacity—including the session key size—without altering any other parameters, making it well suited for real‐world deployment. Because diffusion models allow users to select image resolutions freely, our scheme naturally accommodates different resolutions without additional tuning, whereas methods such as GS \cite{gs_2024} require modifying the diffusion factor, which can be cumbersome.

Under adversarial distortions, we observe a slight decline in bit accuracy as the number of inversion steps increases. We attribute this to complex interactions between certain perturbations, such as brightness changes, and the inversion process.

\begin{table}[htbp]
\centering
\caption{ Robustness evaluation of T2SMark across different inversion steps, shown as TPR/Bit Acc.}
\label{tab:detailed_np}
\begin{tabular}{lccccc}
\toprule
Noise       & 5         & 10        & 25        & 50        & 100       \\
\midrule
None       & 1.000/1.0000 & 1.000/1.0000 & 1.000/1.0000 & 1.000/1.0000 & 1.000/1.0000 \\
JPEG        & 1.000/0.9886 & 1.000/0.9901 & 1.000/0.9904 & 1.000/0.9899 & 1.000/0.9879 \\
RandCr  & 1.000/0.9899 & 1.000/0.9934 & 1.000/0.9899 & 1.000/0.9857 & 1.000/0.9816 \\
RandDr  & 1.000/0.9851 & 1.000/0.9901 & 1.000/0.9873 & 1.000/0.9836 & 1.000/0.9791 \\
Resize      & 1.000/0.9955 & 1.000/0.9963 & 1.000/0.9969 & 1.000/0.9968 & 1.000/0.9967 \\
GauBlur& 1.000/0.9835 & 1.000/0.9890 & 1.000/0.9878 & 1.000/0.9871 & 1.000/0.9867 \\
MedBlur  & 1.000/0.9982 & 1.000/0.9984 & 1.000/0.9981 & 1.000/0.9981 & 1.000/0.9982 \\
GauNoise&0.972/0.8849 & 0.986/0.9018 & 0.976/0.9077 & 0.992/0.9062 & 0.984/0.9034 \\
S\&PNoise        & 0.996/0.9373 & 0.996/0.9482 & 0.998/0.9530 & 0.998/0.9504 & 1.000/0.9535 \\
Brightness  & 0.994/0.9741 & 0.992/0.9803 & 0.990/0.9760 & 0.990/0.9728 & 0.990/0.9698 \\
Adv. (ave)  & 0.996/0.9708 & 0.997/0.9764 & 0.996/0.9764 & 0.998/0.9745 & 0.997/0.9730 \\
\bottomrule
\end{tabular}
\end{table}

\begin{table}[htbp]
  \centering
  \begin{minipage}[b]{0.430\textwidth}
    \centering
    \caption{Bit accuracy of T2SMark watermark extraction across different embedding capacities.}
    \label{tab:detailed_cap}
    \resizebox{\textwidth}{!}{
    \begin{tabular}{lccccc}
      \toprule
      Noise       & 256     & 384     & 512     & 768     & 1024    \\
      \midrule
      None        & 1.0000  & 0.9999  & 0.9999  & 0.9992  & 0.9968  \\
      JPEG        & 0.9901  & 0.9766  & 0.9648  & 0.9381  & 0.9107  \\
      RandCr      & 0.9935  & 0.9814  & 0.9630  & 0.9313  & 0.8936  \\
      RandDr      & 0.9895  & 0.9774  & 0.9488  & 0.9236  & 0.8425  \\
      Resize      & 0.9963  & 0.9900  & 0.9816  & 0.9585  & 0.9332  \\
      GauBlur     & 0.9889  & 0.9700  & 0.9524  & 0.9140  & 0.8816  \\
      MedBlur     & 0.9984  & 0.9939  & 0.9878  & 0.9692  & 0.9482  \\
      GauNoise    & 0.8961  & 0.8691  & 0.8487  & 0.8123  & 0.7744  \\
      S\&PNoise   & 0.9456  & 0.9147  & 0.8960  & 0.8547  & 0.8205  \\
      Brightness  & 0.9803  & 0.9628  & 0.9509  & 0.9287  & 0.9056  \\
      Adv.\ (ave) & 0.9754  & 0.9595  & 0.9438  & 0.9145  & 0.8789  \\
      \bottomrule
    \end{tabular}}
  \end{minipage}%
  \hfill
  \begin{minipage}[b]{0.54\textwidth}
    \centering
    \caption{Robustness performance of T2SMark watermark with different random key sizes, shown as TPR/Bit Acc.}
    \label{tab:detailed_key}
    \resizebox{\textwidth}{!}{
    \begin{tabular}{lcccc}
      \toprule
      Noise       & 8             & 16            & 24            & 32            \\
      \midrule
      None        & 1.000/0.9999  & 1.000/1.0000  & 1.000/1.0000  & 1.000/0.9999  \\
      JPEG        & 1.000/0.9900  & 1.000/0.9901  & 1.000/0.9898  & 1.000/0.9730  \\
      RandCr      & 1.000/0.9930  & 1.000/0.9935  & 1.000/0.9937  & 1.000/0.9718  \\
      RandDr      & 1.000/0.9897  & 1.000/0.9895  & 1.000/0.9880  & 1.000/0.9787  \\
      Resize      & 1.000/0.9966  & 1.000/0.9963  & 1.000/0.9973  & 1.000/0.9930  \\
      GauBlur     & 1.000/0.9868  & 1.000/0.9890  & 1.000/0.9862  & 1.000/0.9760  \\
      MedBlur     & 1.000/0.9980  & 1.000/0.9984  & 1.000/0.9984  & 1.000/0.9982  \\
      GauNoise    & 0.994/0.9113  & 0.994/0.8961  & 0.982/0.8629  & 0.940/0.8172  \\
      S\&PNoise   & 0.998/0.9512  & 0.992/0.9456  & 1.000/0.9340  & 0.998/0.8802  \\
      Brightness  & 0.990/0.9803  & 0.992/0.9803  & 0.986/0.9680  & 0.984/0.9443  \\
      Adv.\ (ave) & 0.998/0.9774  & 0.998/0.9754  & 0.996/0.9687  & 0.991/0.9481  \\
      \bottomrule
    \end{tabular}}
  \end{minipage}
\end{table}

\subsection{Analysis of Computational Overhead}

We conducted a timing comparison for all watermark methods. Experiments were run on a system equipped with an AMD EPYC 7662 Processor and an NVIDIA RTX A6000 GPU. All inversion-based methods utilized FP16 precision and performed 10 inversion steps on an SD v2.1 model \cite{ldm_2022}.

\begin{table}[h!]
\centering
\caption{Timing comparison of various watermarking methods. Embedding and verification times are measured in seconds (s).}
\label{tab:timing_comparison}
\begin{tabular}{lcc}
\toprule
\textbf{Method} & \textbf{Embedding Time (s)} & \textbf{Verification Time (s)} \\
\midrule
dwtDct\cite{dwt2007}      & 0.047 & 0.030 \\
dwtDctSvd\cite{dwt2007}   & 0.113 & 0.065 \\
rivaGan\cite{rivagan2019}     & 1.327 & 1.087 \\
StableSig\cite{stablesig_2023}   & 0.000 & 0.004 \\
TRW\cite{treering_2023}         & 0.001 & 0.437 \\
GS\cite{gs_2024}          & 2.253 & 0.473 \\
PRCW\cite{prcw_2025}        & 0.006 & 0.821 \\
T2SMark     & 0.024 & 0.420 \\
\bottomrule
\end{tabular}
\end{table}

\subsection{Impact of the Two-Stage Framework}

To assess the impact of the two-stage framework, we evaluate T2SMark both with and without it. All experiments use Stable Diffusion v2.1 \cite{ldm_2022} under the same settings as the main paper. For robustness testing, we sample 500 prompts from the Stable-Diffusion-Prompts training split; for generation diversity, we use 1,000 prompts from the same dataset and report LPIPS scores \cite{lpips_2018}. Results are shown in Table~\ref{tab:2s}.

Enabling the two-stage framework reduces adversarial bit accuracy from 0.9868 to 0.9754, since reserving latent dimensions to encrypt the session key cuts redundancy and causes cascading errors—any mistake in decoding the session key invalidates the second-stage decoding. However, it substantially increases generation diversity, confirming its importance for balancing robustness and diversity. In contrast, without two-stage encryption, T2SMark suffers from the same fixed-codeword limitation as Gaussian Shading \cite{gs_2024} and overconcentrates energy in the high-energy tail region, whose position in the latent vector is entirely key-dependent
, significantly reducing diversity.

\begin{table}[h]
  \centering
    \caption{Performance of T2SMark both with and without the two-stage framework.}
  \label{tab:2s}
  \begin{tabular}{lcc}
    \toprule
              & Bit Acc. (Clean/Adv.)    &  Diversity \(\uparrow\)          \\
    \midrule
    w/o two-stage       & 1.0000/0.9868 & 0.5689 \\
    w/ two-stage        & 1.0000/0.9754 & 0.6746 \\
    \bottomrule
    \end{tabular}
\end{table}

\section{Visual Results}

\begin{figure}[t]
  \centering
  \includegraphics[width=0.8\linewidth]{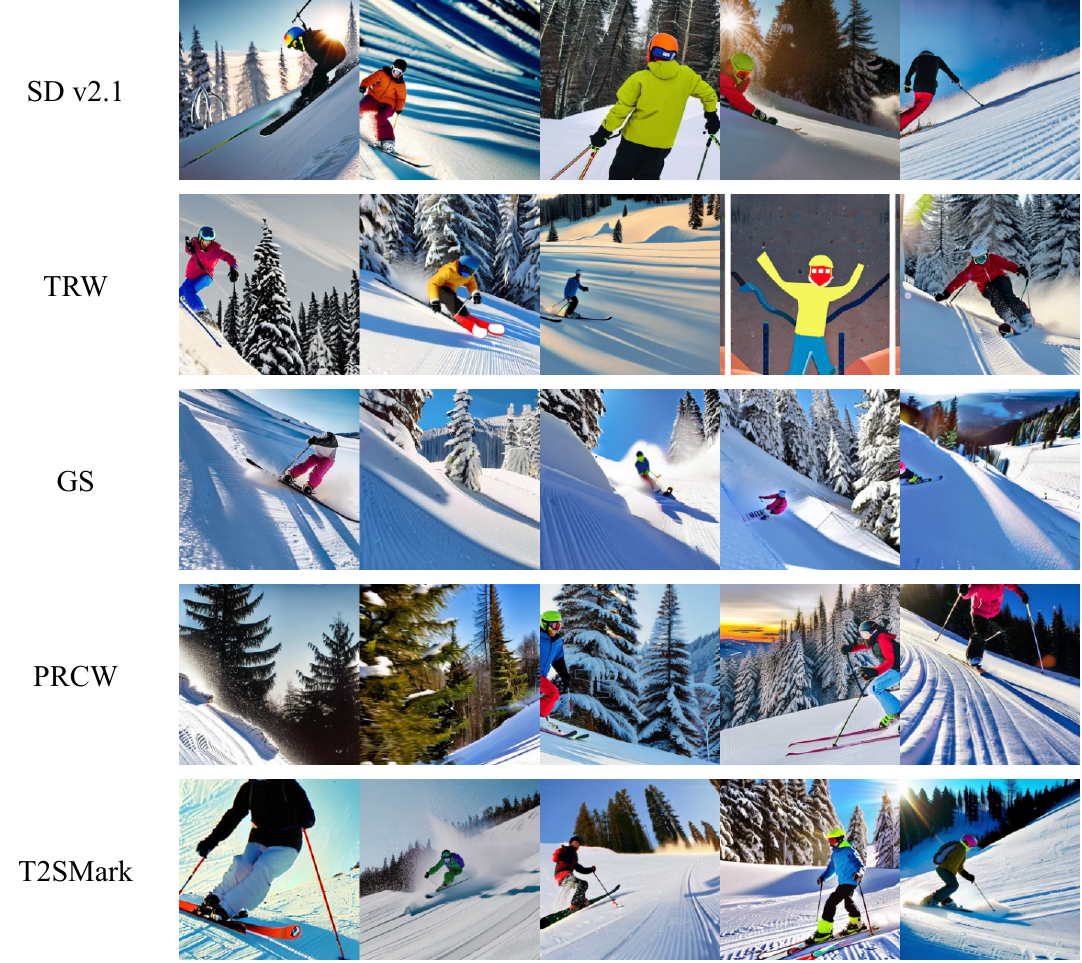}
  \caption{Diversity of images generated by different watermarking methods via SD v2.1 (guidance scale = 7.5, 50 inference steps) on the prompt \textit{“A guy is having fun skiing down the slope of the hill.”}}
  \label{fig:diversity21-1}
\end{figure}

\begin{figure}[t]
  \centering
  \includegraphics[width=0.8\linewidth]{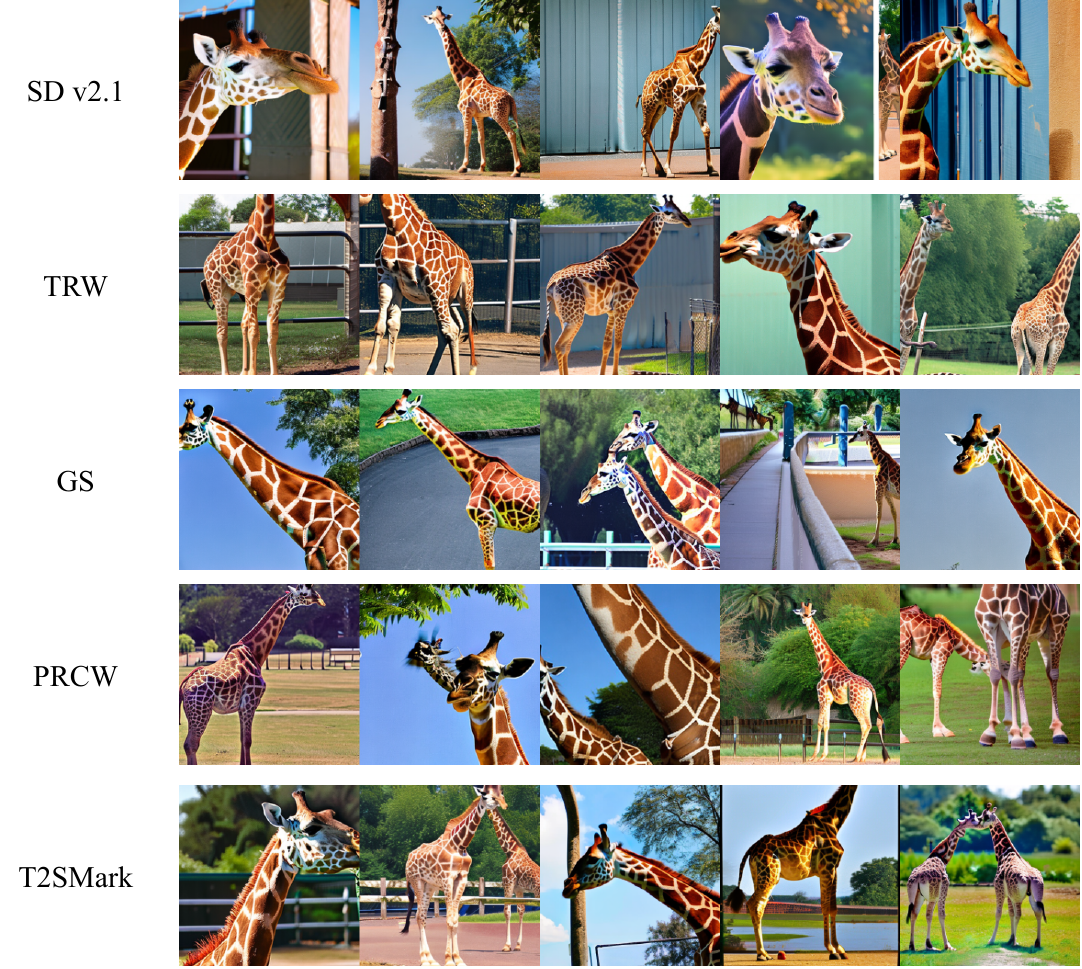}
  \caption{Diversity of images generated by different watermarking methods via SD v2.1 (guidance scale = 7.5, 50 inference steps) on the prompt \textit{“Two giraffes in a zoo enjoy a walk and a snack.”}}
  \label{fig:diversity21-2}
\end{figure}

\begin{figure}[t]
  \centering
  \includegraphics[width=0.8\linewidth]{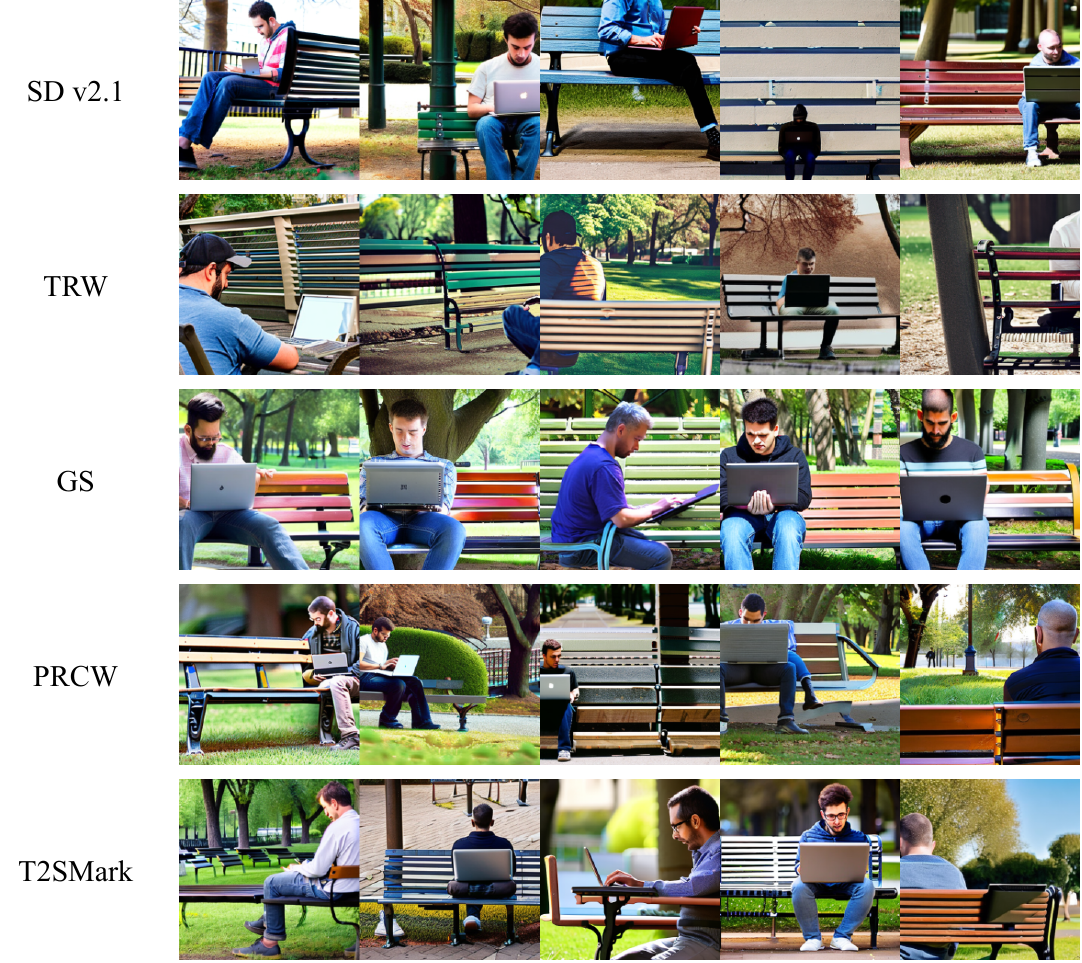}
  \caption{Diversity of images generated by different watermarking methods via SD v2.1 (guidance scale = 7.5, 50 inference steps) on the prompt \textit{“A man uses his computer on a park bench.”}}
  \label{fig:diversity21-3}
\end{figure}

\begin{figure}
    \centering
    \includegraphics[width=0.8\linewidth]{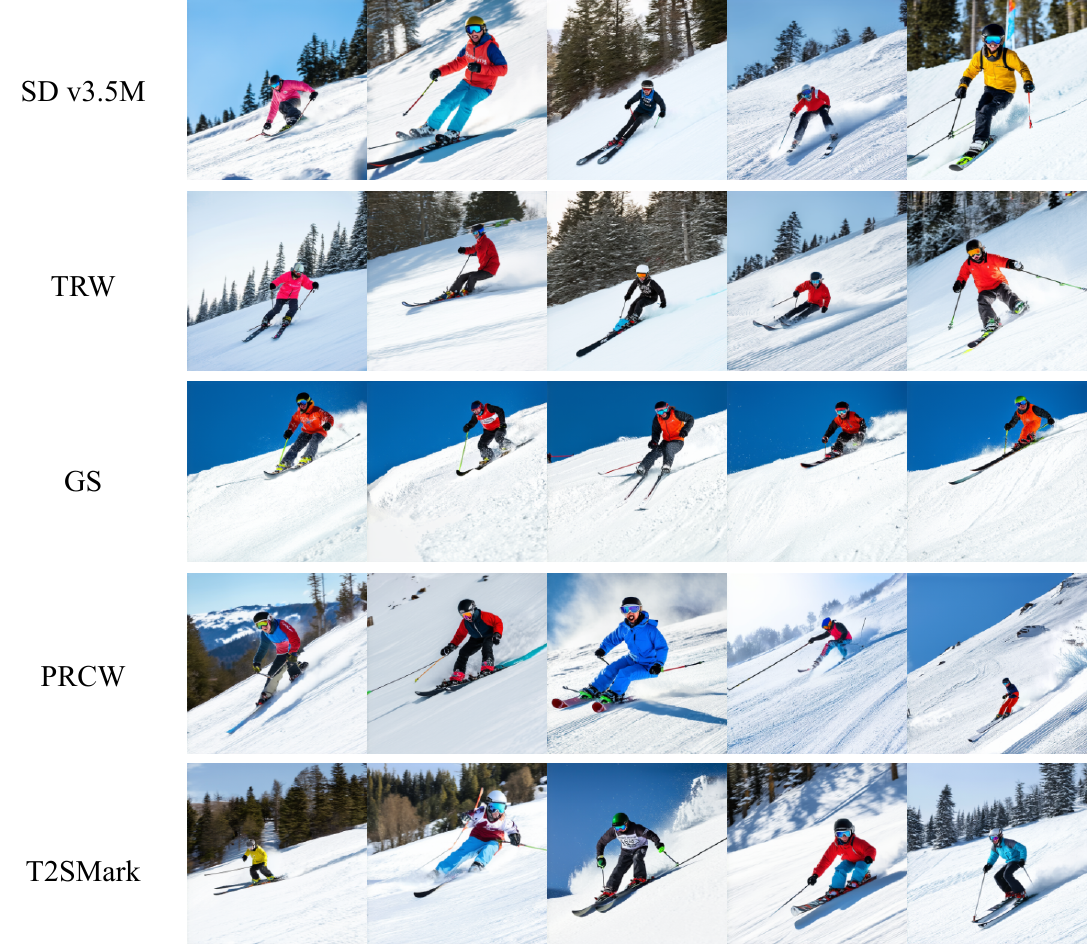}
    \caption{Diversity of images generated by different watermarking methods using SD v3.5M (guidance scale = 4.0, 40 inference steps) on the prompt \textit{“A guy is having fun skiing down the slope of the hill.”}}
    \label{fig:diversity35}
\end{figure}

\paragraph{Generation Diversity.}
Figure \ref{fig:diversity21-1}, \ref{fig:diversity21-2}, \ref{fig:diversity21-3}, and \ref{fig:diversity35} visually compare the generation diversity achieved by various watermarking methods on SD v2.1 and SD v3.5M. Images produced with Gaussian Shading \cite{gs_2024}, for example, display a highly consistent layout, indicating limited diversity.

\clearpage

\clearpage
\newpage
\section*{NeurIPS Paper Checklist}

\begin{enumerate}

\item {\bf Claims}
    \item[] Question: Do the main claims made in the abstract and introduction accurately reflect the paper's contributions and scope?
    \item[] Answer: \answerYes{} 
    \item[] Justification: Section~\ref{sec:intro} presents the main contributions of this paper, which are concisely summarized in both the abstract and the introduction.

\item {\bf Limitations}
    \item[] Question: Does the paper discuss the limitations of the work performed by the authors?
    \item[] Answer: \answerYes{} 
    \item[] Justification: The potential attack vectors and usage limitations of our method are discussed in Section \ref{sec:limitations}.
    \item[] Guidelines:
    \begin{itemize}
        \item The answer NA means that the paper has no limitation while the answer No means that the paper has limitations, but those are not discussed in the paper. 
        \item The authors are encouraged to create a separate "Limitations" section in their paper.
        \item The paper should point out any strong assumptions and how robust the results are to violations of these assumptions (e.g., independence assumptions, noiseless settings, model well-specification, asymptotic approximations only holding locally). The authors should reflect on how these assumptions might be violated in practice and what the implications would be.
        \item The authors should reflect on the scope of the claims made, e.g., if the approach was only tested on a few datasets or with a few runs. In general, empirical results often depend on implicit assumptions, which should be articulated.
        \item The authors should reflect on the factors that influence the p\texttt{elf}ormance of the approach. For example, a facial recognition algorithm may p\texttt{elf}orm poorly when image resolution is low or images are taken in low lighting. Or a speech-to-text system might not be used reliably to provide closed captions for online lectures because it fails to handle technical jargon.
        \item The authors should discuss the computational efficiency of the proposed algorithms and how they scale with dataset size.
        \item If applicable, the authors should discuss possible limitations of their approach to address problems of privacy and fairness.
        \item While the authors might fear that complete honesty about limitations might be used by reviewers as grounds for rejection, a worse outcome might be that reviewers discover limitations that aren't acknowledged in the paper. The authors should use their best judgment and recognize that individual actions in favor of transparency play an important role in developing norms that preserve the integrity of the community. Reviewers will be specifically instructed to not penalize honesty concerning limitations.
    \end{itemize}

\item {\bf Theory assumptions and proofs}
    \item[] Question: For each theoretical result, does the paper provide the full set of assumptions and a complete (and correct) proof?
    \item[] Answer: \answerYes{} 
    \item[] Justification: This paper includes theoretical result, and the assumption and proof are provided in Appendix \ref{sec:ber}.

    \item {\bf Experimental result reproducibility}
    \item[] Question: Does the paper fully disclose all the information needed to reproduce the main experimental results of the paper to the extent that it affects the main claims and/or conclusions of the paper (regardless of whether the code and data are provided or not)?
    \item[] Answer: \answerYes{} 
    \item[] Justification: We include complete experimental details in Section \ref{sec:setting} and Appendix \ref{sec:app_exp_details} to enable reproduction of our results .

\item {\bf Open access to data and code}
    \item[] Question: Does the paper provide open access to the data and code, with sufficient instructions to faithfully reproduce the main experimental results, as described in supplemental material?
    \item[] Answer: \answerYes{} 
    \item[] Justification: Our code is provided in the supplementary material, with detailed usage instructions available in the accompanying README.

\item {\bf Experimental setting/details}
    \item[] Question: Does the paper specify all the training and test details (e.g., data splits, hyperparameters, how they were chosen, type of optimizer, etc.) necessary to understand the results?
    \item[] Answer: \answerYes{} 
    \item[] Justification: We provide detailed settings in Section \ref{sec:setting} and Appendix \ref{sec:app_exp_details}.

\item {\bf Experiment statistical significance}
    \item[] Question: Does the paper report error bars suitably and correctly defined or other appropriate information about the statistical significance of the experiments?
    \item[] Answer: \answerYes{} 
    \item[] Justification: We report the means and standard errors for experimental results in Tables \ref{tab:results21}, \ref{tab:results35}, and \ref{tab:results21_dataset}.

\item {\bf Experiments compute resources}
    \item[] Question: For each experiment, does the paper provide sufficient information on the computer resources (type of compute workers, memory, time of execution) needed to reproduce the experiments?
    \item[] Answer: \answerYes{} 
    \item[] Justification: The GPU used to conduct our experiments is specified in Section~\ref{sec:setting}.
    
\item {\bf Code of ethics}
    \item[] Question: Does the research conducted in the paper conform, in every respect, with the NeurIPS Code of Ethics \url{https://neurips.cc/public/EthicsGuidelines}?
    \item[] Answer: \answerYes{} 
    \item[] Justification: The research conducted in the paper conforms with the NeurIPS Code of Ethics.

\item {\bf Broader impacts}
    \item[] Question: Does the paper discuss both potential positive societal impacts and negative societal impacts of the work performed?
    \item[] Answer: \answerNA{} 
    \item[] Justification: The answer NA means that there is no societal impact of the work performed.
    
\item {\bf Safeguards}
    \item[] Question: Does the paper describe safeguards that have been put in place for responsible release of data or models that have a high risk for misuse (e.g., pretrained language models, image generators, or scraped datasets)?
    \item[] Answer: \answerNA{} 
    \item[] Justification: The answer NA means that the paper poses no such risks.

\item {\bf Licenses for existing assets}
    \item[] Question: Are the creators or original owners of assets (e.g., code, data, models), used in the paper, properly credited and are the license and terms of use explicitly mentioned and properly respected?
    \item[] Answer: \answerNA{} 
    \item[] Justification: The answer NA means that the paper does not use existing assets

\item {\bf New assets}
    \item[] Question: Are new assets introduced in the paper well documented and is the documentation provided alongside the assets?
    \item[] Answer: \answerNA{} 
    \item[] Justification: The answer NA means that the paper does not release new assets.

\item {\bf Crowdsourcing and research with human subjects}
    \item[] Question: For crowdsourcing experiments and research with human subjects, does the paper include the full text of instructions given to participants and screenshots, if applicable, as well as details about compensation (if any)? 
    \item[] Answer: \answerNA{} 
    \item[] Justification: The answer NA means that the paper does not involve crowdsourcing nor research with human subjects.

\item {\bf Institutional review board (IRB) approvals or equivalent for research with human subjects}
    \item[] Question: Does the paper describe potential risks incurred by study participants, whether such risks were disclosed to the subjects, and whether Institutional Review Board (IRB) approvals (or an equivalent approval/review based on the requirements of your country or institution) were obtained?
    \item[] Answer: \answerNA{} 
    \item[] Justification:  The answer NA means that the paper does not involve crowdsourcing nor research with human subjects.

\item {\bf Declaration of LLM usage}
    \item[] Question: Does the paper describe the usage of LLMs if it is an important, original, or non-standard component of the core methods in this research? Note that if the LLM is used only for writing, editing, or formatting purposes and does not impact the core methodology, scientific rigorousness, or originality of the research, declaration is not required.
    \item[] Answer: \answerNA{} 
    \item[] Justification: The answer NA means that the core method development in this research does not involve LLMs as any important, original, or non-standard components.

\end{enumerate}

\end{document}